\documentclass{article}

\PassOptionsToPackage{numbers}{natbib}
\usepackage[final,main]{neurips_2026}

\usepackage[utf8]{inputenc}
\usepackage[T1]{fontenc}
\usepackage{hyperref}
\usepackage{url}
\usepackage{booktabs}
\usepackage{amsfonts}
\usepackage{amssymb}
\usepackage{amsmath}
\usepackage{nicefrac}
\usepackage{microtype}
\usepackage{xcolor}
\usepackage{graphicx}
\usepackage{tcolorbox}
\usepackage{listings}
\usepackage{placeins}
\title{When Trackers Fail: VLM-Guided Verification and Recovery for Robust Video Object Segmentation}

\author{%
  Valay Bundele$^{1}$\quad Susmit Agrawal$^{1}$\quad Mehran Hosseinzadeh$^{1}$\\
  \textbf{The Nam Nguyen$^{2}$\quad Hendrik P.\ A.\ Lensch$^{1}$}\\
  $^{1}$University of T\"ubingen, Germany\quad $^{2}$Woven by Toyota, Japan\\
  \texttt{\{valay.bundele,mehran.hosseinzadeh,hendrik.lensch\}@uni-tuebingen.de}\\
  \texttt{susmit.agrawal@bethgelab.org}
}

\begin{document}

\maketitle

\begin{abstract}
Recent trackers built on foundation models such as the Segment Anything Model (SAM) achieve strong performance by propagating object masks through memory. However, this strength is also a key source of failure: under occlusion, reappearance, or distractors, an incorrect prediction can enter the tracking loop and be propagated over time, resulting in persistent identity drift. Existing approaches seek to prevent such failures through improved memory and temporal modeling, but once a failure occurs, the tracker has no explicit mechanism to recognize or correct it. We address this gap with Dual-Agent, a failure-aware tracking framework that complements propagation with explicit verification and recovery. A vision-language verification agent first reasons over spatio-temporal context to determine whether the tracker is still following the target. When a failure is detected, a correction agent uses temporal context to recover the target before tracking continues. The correction agent is trained on real SAM3 failures, augmented with targeted identity-switch and distractor perturbations that expose it to challenging ambiguities during recovery. On LVOS-v2 and SecVOS, Dual-Agent improves SAM3 by 1.2 and 2.7 $J\&F$, respectively. Importantly, the proposed framework transfers without retraining to SAM2.1, SeC, and Cutie, as well as to long-term video object tracking benchmarks including LaSOT-ext, TNL2K, and VOT-LT2022. Our results show that explicitly reasoning about when tracking fails and how to recover provides a complementary path to robust tracking beyond improving propagation alone. Project page: \href{https://valaybundele.github.io/Dual-Agent}{\texttt{valaybundele.github.io/Dual-Agent}}.
\end{abstract}

    

\section{Introduction}
\begin{figure}[t]
    \centering
    
    \includegraphics[width=0.9\textwidth]{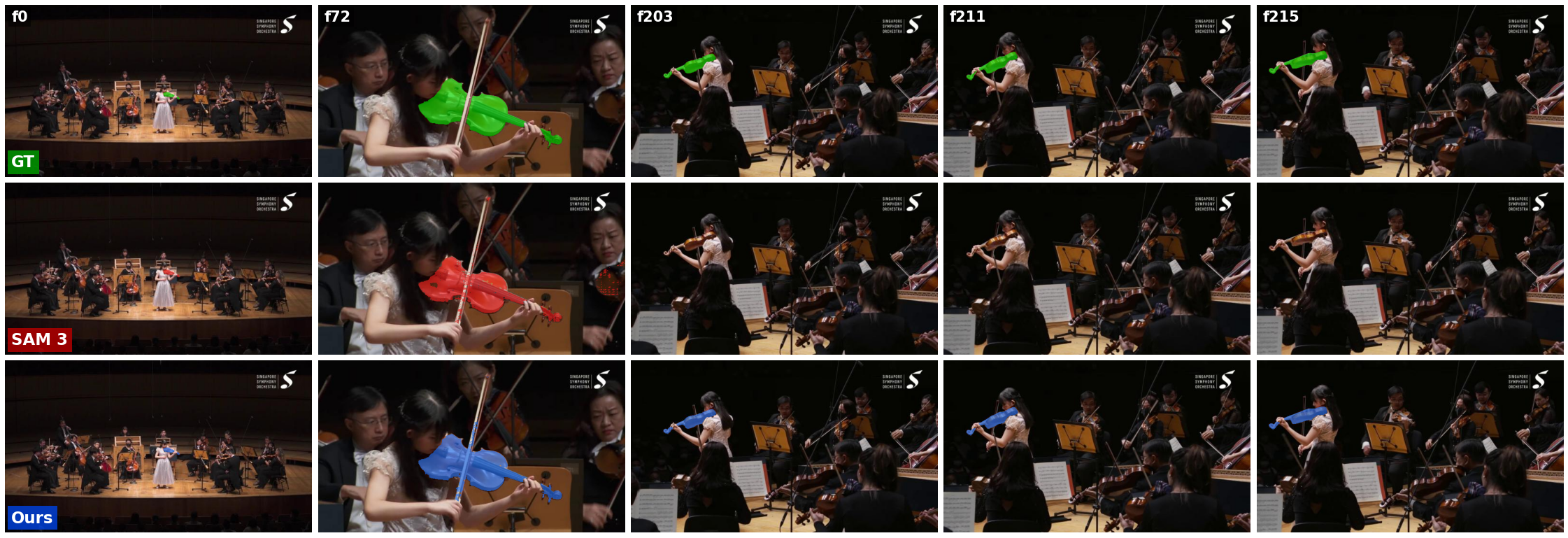}
    \caption{Our proposed Dual-Agent framework recovers tracking errors made by the tracker. Rows show ground truth (top), the SAM3 baseline (middle), and our Dual-Agent framework (bottom).}
    \label{fig:teaser}
\end{figure}

Segmentation-based tracking has recently undergone a paradigm shift with the emergence of foundation models such as Segment Anything Model and its video extensions (e.g., SAM2~\citep{ravi2024sam} and SAM3~\citep{carion2025sam}), which enable promptable mask propagation across frames. These approaches achieve strong performance by combining powerful visual representations with streaming memory, and have quickly become the dominant paradigm for video object segmentation and tracking. However, despite their impressive generalization, they remain fundamentally brittle in challenging scenarios involving occlusion, reappearance, and distractors. In such cases, trackers often drift toward visually similar objects or fail to recover the target after disappearance, revealing a lack of explicit scene understanding and failure awareness (see Figure \ref{fig:teaser}).

A growing body of recent work has focused on improving SAM-based trackers by refining temporal modeling and memory design. For example, SAMURAI~\citep{yang2024samurai} introduces motion-aware memory selection to incorporate temporal motion cues and mitigate error propagation. Similarly, DAM4SAM~\citep{videnovic2025distractor} proposes a distractor-aware memory mechanism that explicitly models confusing objects and improves robustness under occlusion. Other extensions explore long-term memory and improved memory management strategies ~\citep{ding2025sam2long}. More recently, SeC~\citep{zhang2025sec} introduces progressive concept construction to improve robustness under complex scene changes by incorporating higher-level semantic priors into segmentation. While these approaches achieve substantial gains, they all share a common design principle: \textit{tracking is performed via recursive propagation of previous predictions}. As a result, failures caused by occlusion, distractors, or identity ambiguity are not explicitly identified; incorrect predictions are simply propagated forward, often leading to persistent drift.

In parallel, recent advances in vision-language models (VLMs) have demonstrated strong capabilities in semantic understanding and contextual reasoning~\citep{bai2025qwen3, comanici2025gemini}. Motivated by these properties, a growing line of work in referring and reasoning-based video object segmentation incorporates VLMs to identify or disambiguate target objects based on language or contextual cues~\citep{wu2022language, cuttano2025samwise, li2025refsam}. However, in these approaches, VLMs are primarily used for object grounding or selection, while the actual tracking remains handled by a conventional propagation-based model. Thus, while they improve what to track, they do not address a critical question: \textit{whether the tracker remains correct over time}. In particular, they lack an explicit mechanism for detecting and correcting failures during tracking.

In this work, we argue that the central limitation of modern tracking systems is not the lack of better propagation, but \textit{the absence of failure-aware reasoning}. Existing approaches implicitly assume that predictions remain valid over time, and thus focus on improving temporal consistency. However, as evidenced by recent SAM-based extensions, even sophisticated memory and motion modeling cannot fully prevent drift once failures occur, since they are never explicitly diagnosed or corrected ~\citep{yang2024samurai, videnovic2025distractor}.

To address this, we propose a fundamentally different formulation that introduces explicit reasoning into the tracking loop. 
Rather than relying on propagation alone, our key idea is to complement it with two capabilities: (i) verifying whether the current prediction remains correct, and (ii) recovering the target when it does not. We instantiate this idea with \textit{Dual-Agent}, a framework comprising a verification agent and a correction agent alongside the underlying propagator. The verification agent is a vision-language reasoning model that operates over spatio-temporal context to assess whether the current prediction still corresponds to the target. Rather than relying solely on heuristic confidence scores, it reasons about object identity, visibility, and consistency across frames, enabling it to recognize identity switches, missed detections, and false positives during occlusion and reappearance.

When a failure is detected, the correction agent is selectively activated to recover the target using temporal context before propagation continues. Effective recovery, however, requires resolving precisely the identity ambiguities that cause tracking to fail. We therefore train the correction agent on failures mined from SAM3 and augment them with targeted identity-switch and distractor-centric perturbations. These perturbations expose the model to challenging target–distractor ambiguities and encourage it to reason about target identity across time rather than relying on local appearance alone. The separation of verification from correction allows the framework to intervene specifically when the propagated prediction becomes unreliable, while otherwise retaining the underlying tracker. 

We evaluate Dual-Agent across video object segmentation and long-term object tracking benchmarks. On LVOS-v2 and SecVOS, Dual-Agent improves SAM3 by 1.2 and 2.7 $J\&F$, respectively, while on LaSOT-ext, TNL2K, and VOT-LT2022, it yields gains of 1-2\% in long-term object tracking. Beyond SAM3, the proposed framework transfers without retraining to other propagators, including SAM2.1, SeC, and Cutie, demonstrating that the verification and recovery mechanisms generalize across different propagation-based trackers.

Overall, our Dual-Agent framework with SAM3 as the propagator reframes segmentation-based tracking from improving propagation to explicitly modeling failure detection and recovery. By integrating structured reasoning into the tracking loop, our approach addresses a fundamental limitation shared by both SAM-based trackers and prior VLM-assisted methods. More broadly, our results suggest that incorporating reasoning into perception systems is a promising direction for building robust, self-correcting vision models capable of operating in complex, real-world environments.

\section{Related Work}



\paragraph{Memory-based video object segmentation.}
Modern semi-supervised VOS systems are dominated by memory-based propagation. XMem~\citep{cheng2022xmem} established the importance of explicit long-term and short-term memory for stable mask propagation, while later methods such as Cutie~\citep{cheng2024putting} improved object-centric memory reading and multi-object robustness. Related work has further explored stronger retrieval and identity modeling through object-centric matching and retrieval mechanisms~\citep{wang2023look, yan2023universal}, as well as unified prompt-driven frameworks such as TarViS~\citep{athar2023tarvis} and UniVS~\citep{li2024univs}. Foundation-model-based trackers including SAM2~\citep{ravi2024sam}, SAM2Long~\citep{ding2025sam2long}, SAMURAI~\citep{yang2024samurai}, DAM4SAM~\citep{videnovic2025distractor}, and SAM3~\citep{carion2025sam} further improve propagation through motion-aware memory selection, distractor-aware memory writing, and long-video memory management. However, these approaches still rely on recursive propagation of previous predictions and do not explicitly reason about whether the tracker has already failed. Once drift occurs, corrupted state can be propagated and reinforced over time. In contrast, our approach introduces an explicit reliability layer on top of propagation that detects and selectively corrects failures before they dominate future tracking.

\paragraph{VLMs for video understanding and tracking.}
VLMs have recently demonstrated strong capabilities in semantic grounding, contextual disambiguation, and visual reasoning. Language-guided VOS methods such as ReferFormer~\citep{wu2022language}, RefSAM~\citep{li2025refsam}, and SAMWISE~\citep{cuttano2025samwise} use language prompts to specify target objects and adapt the propagation pipeline accordingly. More recent reasoning-oriented approaches including VISA~\citep{yan2024visa}, ViLLa~\citep{zheng2407villa}, CoT-RVS~\citep{kao2025cot}, VideoLISA~\citep{bai2024one}, and VideoSeg-R1~\citep{xu2026videoseg} incorporate higher-level semantic reasoning into video segmentation. Training-free agentic methods such as AL-Ref-SAM2~\citep{huang2025unleashing}, AgentRVOS~\citep{jin2026agentrvos}, and Refer-Agent~\citep{jiang2026refer} similarly combine large multimodal models with SAM-style propagation, while SeC~\citep{zhang2025sec} injects concept-level priors directly into the backbone to improve robustness under complex scene changes. However, these methods primarily use reasoning to improve target grounding, representation, or segmentation quality during propagation. Unlike SeC that probes and refines intermediate tracker representations, our framework does not modify the propagator internals or memory features. In contrast, we introduce explicit failure-aware reasoning into the tracking loop itself. Our framework reasons over tracker-generated, potentially corrupted temporal context to determine whether the current track remains reliable, and selectively triggers recovery when failures such as identity drift or missed reappearance are detected.


\section{Methodology}

\subsection{Problem Formulation}
 
 Given a video $\{I_t\}_{t=1}^{T}$ and an initial mask prompt on the first frame, a tracker produces a sequence of masks $\{\hat{M}_t\}_{t=1}^{T}$. In SAM3-style propagation, the estimate at frame $t$ is produced from the current image and an internal temporal state,
namely $\hat{M}_t = \mathcal{T}(I_t, \mathcal{S}_{t-1})$,
where $\mathcal{S}_{t-1}$ summarizes memory from previous frames. The next state $\mathcal{S}_t$ is then updated from the current prediction. This recursive formulation is powerful, but it creates a failure mode that is especially severe under occlusion and distractors: once the tracker predicts wrong object or loses target, \textit{the error is fed back into memory and propagated forward}. 

Our goal is to augment this propagation with explicit failure awareness. Rather than trusting every prediction equally, we seek a mechanism that can (i) determine when the tracker is likely wrong, and (ii) selectively recover the target when failure occurs. We therefore keep SAM3 as the propagator and target the regime in which propagation becomes unreliable. This yields a different view of VOS from most prior work: the central problem is not only how to propagate masks better, but \textit{how to regulate propagation once its internal state becomes untrustworthy.}

The resulting framework is guided by four design principles that also summarize the main technical contributions of the paper. First, we use an \emph{asymmetric decomposition} of the problem into verification and correction, because deciding \emph{whether} the track is wrong and deciding \emph{where} the target is should not be treated as the same task. Second, we use a \emph{tracker-aware reasoning interface}: the verifier reasons over a prompt that exposes both trusted exemplars and tracker-produced potentially corrupted context, rather than over clean oracle trajectories. Third, we use a \emph{state-safe intervention policy} combining sparse triggers and commit gating, motivated by the fact that an incorrect correction is costly because it rewrites future tracker memory. Fourth, we use \emph{failure-oriented supervision} for recovery, including identity-switch and distractor-centric perturbations that explicitly train the corrector on the kinds of ambiguity that cause SAM-style propagators to fail.

\begin{figure}[t]
    \centering
    \includegraphics[width=0.8\textwidth]{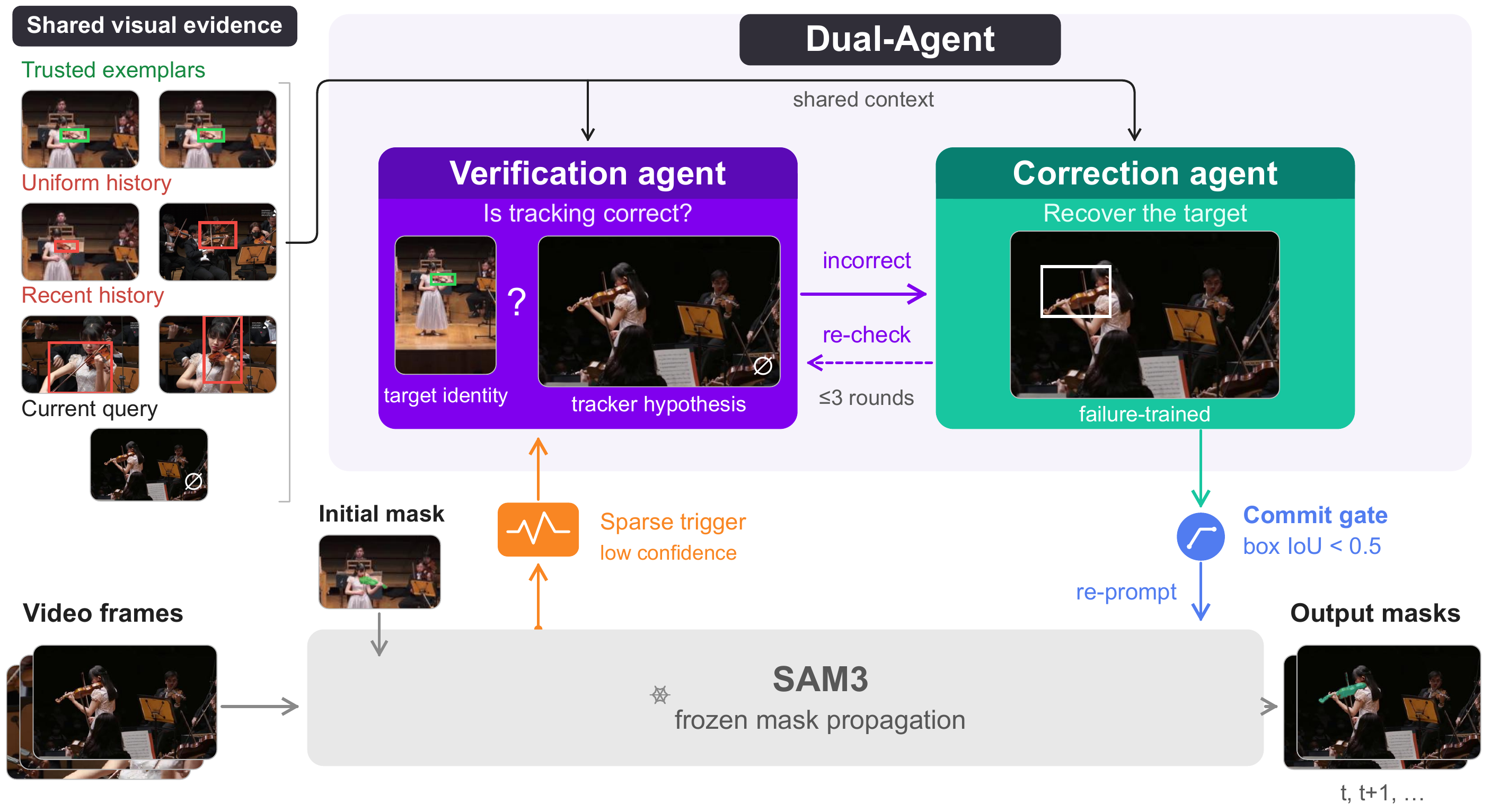}
    \caption{\textbf{Dual-Agent adds a verification and recovery layer above frozen SAM3.} Shared evidence contains exemplars, tracker history and a query. Sparse triggers and confidence gating invoke verification; an incorrect decision activates the failure-trained corrector. Intermediate box proposals may be rechecked for up to three correction rounds. A separate commit gate uses box IoU below 0.5 when both boxes exist. Committed prompts update SAM3; otherwise tracking continues.}
    \label{fig:pipeline}
\end{figure}
\subsection{Dual-Agent Framework}

 We formulate our approach as a Dual-Agent framework that uses SAM3 as propagator and adds two auxiliary agents on top: a verification agent $\mathcal{A}$ and a correction agent $\mathcal{C}$ (Figure \ref{fig:pipeline}). The verifier $\mathcal{A}$ decides whether the current tracked hypothesis is trustworthy; the corrector $\mathcal{C}$ is invoked only when $\mathcal{A}$ predicts failure. 
The dual-agent decomposition keeps these objectives separate and turns reasoning into an explicit control layer over propagation rather than an internal modification of the propagator.

Let $I_t$ denote the video frame at time $t$, $\mathcal{S}_t$ be the internal SAM3 memory state after processing frame $t$, and $\hat{M}_t=\mathcal{T}(I_t,\mathcal{S}_{t-1})$ the propagated mask at time $t$. We convert $\hat{M}_t$ to a box hypothesis $\hat{b}_t=\mathcal{B}(\hat{M}_t)$ and denote the SAM3 confidence score by $s_t \in [0,1]$. The controller is evaluated only at a sparse trigger set $\mathcal{Q}\subset\{1,\ldots,T\}$. At each $t\in\mathcal{Q}$, we build a visual prompt $\mathcal{P}_t$ from tracker history and query frame $I_t$, and apply $
v_t=\mathcal{A}(\mathcal{P}_t)\in\{0,1\},
$
where $\mathcal{A}$ denotes the verifier and $v_t=1$ refers to \texttt{incorrect}. When $v_t=1$, the correction agent $\mathcal{C}$ predicts a visibility variable $\tilde{o}_t\in\{0,1\}$ and a corrected box $\tilde{b}_t\in\mathbb{R}^4$:
$
(\tilde{o}_t,\tilde{b}_t)=\mathcal{C}(\mathcal{P}_t).
$
The tracker state is overwritten only if the commit rule is satisfied. Hence the final output box $b_t^\star$ obeys
\begin{equation}
b_t^\star=
\begin{cases}
\tilde{b}_t, & \text{if } t\in\mathcal{Q},\, v_t=1,\, \Gamma_t=1,\, \tilde{o}_t=1,\\
\emptyset, & \text{if } t\in\mathcal{Q},\, v_t=1,\, \Gamma_t=1,\, \tilde{o}_t=0,\\
\hat{b}_t, & \text{otherwise},
\end{cases}
\end{equation}
where $\Gamma_t$ is the commit gate defined in Section~3.3. If a correction is committed, SAM3 is re-prompted at frame $t$ and propagation resumes from the corrected state rather than from the original prediction. The system therefore intervenes through sparse state resets rather than dense post-hoc mask refinement. In SAM-style trackers, the main cost of an error is not only that one frame is wrong, but that the wrong state is propagated into future frames.

\subsubsection{Verification Agent}

The verifier's goal is to check if the target is being correctly tracked in the current frame given the past target trajectory. For a query time $t$, let
$\mathcal{P}_t=\big(E_t,\,U_t,\,R_t,\,Q_t\big)$ denote the visual prompt, where $E_t$ contains exemplar frames, $U_t$ contains uniformly sampled past context, $R_t$ contains recent context, and $Q_t$ is the query frame. In the main configuration, $|E_t|=2$, $|U_t|=2$, and $|R_t|=2$, yielding a 7-image prompt in total. The exemplars are the earliest visible appearances of the target. The uniform context is sampled from the prior trajectory, while the recent context is sampled with a fixed stride from the immediate past. 

Each frame in $\mathcal{P}_t$ carries a bounding box overlay that exposes the tracker state to the verifier. Exemplar images show ground-truth boxes in green. Context images show tracker-produced boxes in red. The query frame similarly shows the current tracker hypothesis $\hat{b}_t$ in red when present. These context boxes are not oracle boxes: the verifier observes the same possibly corrupted trajectory that the deployed tracker has produced so far. Formally, the verifier computes $\mathcal{A}(\mathcal{P}_t)=v_t$,
where $v_t$ is the binary correctness label. We instantiate $\mathcal{A}$ with Qwen3-VL-8B-Thinking~\citep{bai2025qwen3} and use it zero-shot with a structured assessment prompt (see Appendix D). The prompt asks the model to first locate the target instance in the query image and only then judge whether the red tracker box is on the correct object. This ordering promotes identity-level reasoning rather than local box validation, which is critical when dominant failures are due to semantic swaps.

To reduce single-sample variance, we optionally use repeated verifier calls on temporally shifted query points and aggregate them by majority vote. If the vote count is $K$, the fire decision becomes
$
\bar{v}_t = \mathbf{1}\!\left[\sum_{k=1}^{K} v_t^{(k)} > \frac{K}{2}\right].
$
 This voting mechanism is especially useful in borderline occlusion and reappearance cases, where a single reasoning trace may be unstable.

\subsubsection{Correction Agent}

The correction agent $\mathcal{C}$ uses the same prompt layout as $\mathcal{A}$ but solves a different task: \textit{conditional re-localization under tracker failure}s. Rather than asking for a binary correctness judgment, we ask the model to predict whether the target is visible in the query frame and, if so, to emit a corrected 2D box. The model outputs $
\mathcal{C}(\mathcal{P}_t)=\big(\tilde{o}_t,\tilde{b}^{\mathrm{norm}}_t\big),
$
where $\tilde{b}^{\mathrm{norm}}_t\in[0,1000]^4$ follows the Qwen-VL normalized coordinate convention.

We decouple correction from verification rather than asking the same model to both detect and repair failures. The corrector can then specialize to the error distribution induced by SAM3: identity drift, missed visible targets, false positives during occlusion, and ambiguous reappearance. The verifier is optimized for deciding \emph{whether} intervention is warranted, while the corrector is optimized for deciding \emph{where} the target should be. In our setup, $\mathcal{C}$ is Qwen3-VL-4B-Instruct~\citep{bai2025qwen3} fine-tuned for structured grounded output, which is substantially lighter than the verifier and better suited to repeated localization. The framework therefore allocates semantic reasoning, recovery, and propagation to different roles according to their function rather than forcing a single model to absorb all three.

The implementation also supports iterative correction. After a candidate correction is produced, the verifier may re-evaluate the proposed box; if it is rejected, the corrector can be called again with the rejected box as additional context. Denoting the $r$-th correction by $\tilde{b}^{(r)}_t$, this defines a short recovery chain $
\tilde{b}^{(r)}_t = \mathcal{C}\!\left(\mathcal{P}_t,\tilde{b}^{(r-1)}_t\right), r=1,\ldots,R
$. Iterative refinement is particularly useful in challenging recovery scenarios where a single correction step may fail to fully recover.

\subsection{Selective Triggering and Gating}

Applying a VLM at every frame is both computationally expensive and potentially harmful, since unnecessary corrections can overwrite a valid tracker state. We therefore expose the agents through a sparse trigger process and require additional gates before the corrector can modify the tracker state.


\paragraph{Trigger policy.}
We define a sparse trigger set $\mathcal{Q}$ using three cues from SAM3: occlusion, reappearance, and instability in the tracker's own confidence. The first two are event-based. During occlusion, we poll every $M$ frames, while reappearance triggers are delayed by at least $N$ frames to avoid unstable transient predictions. The third cue is an effective-IoU-variance trigger designed to catch emerging drift before the tracker has fully failed. Recall that SAM3 produces a scalar confidence $s_t$ from its objectness and mask-quality estimates. Rather than thresholding $s_t$ directly, we monitor its short-horizon instability. Let
$
\mu_t = \frac{1}{W}\sum_{i=0}^{W-1} s_{t-i},
\qquad
\sigma_t^2 = \frac{1}{W}\sum_{i=0}^{W-1}(s_{t-i}-\mu_t)^2
$
denote the mean and variance of the recent confidence window. We trigger the verifier when $\sigma_t^2$ exceeds a threshold for a small number of consecutive steps. Intuitively, healthy propagation tends to produce a stable confidence trace, whereas identity swaps or partial occlusion often induce oscillatory confidence even before a hard disappearance event is declared. The variance trigger therefore acts as an early-warning signal complementing the more explicit occlusion and reappearance triggers.  

\paragraph{Commit gate.}
At triggered frames, we bypass the verifier when SAM3 is already confident. Specifically, if $
s_t > \tau_{\mathrm{conf}},
$
we directly accept the propagated prediction. This focuses reasoning on uncertain regimes where memory corruption is most likely.
When the verifier predicts failure, the corrector proposes a visibility decision $\tilde{o}_t$ and a corrected box $\tilde{b}_t$. We commit the correction only if it differs substantially from the current tracker hypothesis. When both boxes are present, we compute $
d_t = \mathrm{IoU}(\tilde{b}_t,\hat{b}_t),
$
and overwrite the tracker state only if $d_t < 0.5$. This prevents unnecessary memory rewrites from near-identical corrections. Overall, the controller enters correction mode only when
$s_t < \tau_{\mathrm{conf}},
\bar{v}_t = 1,t \in \mathcal{Q}
$
and commits the correction only if
$\mathrm{IoU}(\tilde{b}_t,\hat{b}_t) < 0.5
$.
The resulting commit gate is
\begin{equation}
\Gamma_t=
\mathbf{1}[t\in\mathcal{Q}]
\mathbf{1}[s_t<\tau_{\mathrm{conf}}]
\mathbf{1}[\bar{v}_t=1]
\mathbf{1}[d_t<0.5],
\end{equation}
with the final factor set to 1 when the corrector predicts absence.

\subsection{Failure-Oriented Training Data}
\label{sec:training_data}

\paragraph{Data collection.}
The correction agent is trained on data mined from SAM3 propagation trajectories rather than oracle trajectories, so that supervision matches the failure distribution encountered at test time. We initialize SAM3 with the official SA-V masklet at each object's first-appearance frame and propagate to the end of the video on \texttt{sav\_000}--\texttt{sav\_005} \cite{ravi2024sam} (5{,}145 sequences, 16{,}045 objects). For each annotated frame, we record SAM3's prediction, confidence, IoU against ground truth, and a set of visually similar distractor candidates, producing a structured per-object trace.

\paragraph{Sample generation.}
Query frames are selected with a failure-oriented schedule that oversamples frames where SAM3 disagrees with ground truth, while retaining a small fraction of high-IoU frames. For each query, we render the same seven-image prompt used at test time: two early exemplar frames, two uniformly sampled history frames, two recent-context frames, and the query frame itself. The target is a structured visibility decision and ground-truth box. To expose the corrector to identity ambiguity, we additionally apply an identity-switch perturbation with a certain probability in which one or two recent context boxes, and sometimes the query box, are replaced by mined distractor boxes from the same frame. The resulting corpus contains 79{,}457 samples, split sequence-stratified into 75{,}341 train and 4{,}116 validation samples. More details are in the supplementary.

\subsection{Training Objective and Inference}

The corrector is obtained by supervised fine-tuning Qwen3-VL-4B-Instruct with LoRA. Given a training sample $(\mathcal{P}_t,y_t)$, where $y_t$ is a structured JSON answer containing visibility and a normalized box, we optimize the standard autoregressive objective
$
\mathcal{L}_{\mathrm{SFT}}
=-\sum_{n=1}^{N_t}\log p_\theta\!\left(y_{t,n}\mid y_{t,<n},\mathcal{P}_t\right).
$

The full test-time procedure is summarized as follows. SAM3 is initialized with a mask prompt at the object's first appearance and propagates frame by frame while maintaining its internal memory state $\mathcal{S}_{t-1}$. At each triggered frame $t \in \mathcal{Q}$, we construct the visual prompt $\mathcal{P}_t$ and query the verifier unless the confidence short-circuit accepts the current prediction. If the verifier judges the track to be correct, we retain $\hat{b}_t$ and continue propagation. Otherwise, verification is repeated at frames $t+4$ and $t+8$. If a majority of the three verification decisions indicates an incorrect track, the corrector is invoked. The corrected prediction is then committed only if the commit gate is satisfied, in which case the tracker state at frame $t$ is overwritten with the corrected state.
Equivalently, the memory update can be written as
\begin{equation}
\mathcal{S}_t=
\begin{cases}
\mathcal{U}\!\left(\mathcal{S}_{t-1}, I_t, b_t^\star\right), & \Gamma_t=1,\\
\mathcal{U}\!\left(\mathcal{S}_{t-1}, I_t, \hat{b}_t\right), & \Gamma_t=0,
\end{cases}
\end{equation}
where $\mathcal{U}$ denotes the SAM3 state update induced by the accepted prompt at frame $t$. The key point is that the controller does not merely replace the output on one frame; \textit{it changes the future trajectory by altering the memory that subsequent frames will read from}. This is why selective triggering and conservative commitment are central to the method: one incorrect correction can pollute many future frames, while one correct reset can recover a long segment after drift.
\section{Experiments}

\subsection{Experimental Setup}
We evaluate Dual-Agent on five benchmarks covering video object segmentation (VOS) and long-term object tracking (VOT). For VOS, we use LVOS-v2 valid~\citep{hong2025lvos}, comprising 140 sequences and 239 objects with dense per-frame annotations, and SecVOS~\citep{zhang2025sec}, comprising 160 sequences and 182 objects with an emphasis on scene changes and multi-shot discontinuities. We follow the scene-stratified evaluation protocol of SeC for SecVOS and report the standard VOS metrics: region similarity ($\mathcal{J}$), contour accuracy ($\mathcal{F}$), and their mean ($\mathcal{J}\&\mathcal{F}$).
For VOT, we evaluate on LaSOT-ext~\citep{fan2021lasot} (150 sequences), TNL2K~\citep{wang2021tnl2k} (700 sequences), and VOT-LT2022~\citep{kristan2022vot} (50 sequences). We report AUC, precision at 20 pixels, and normalized precision at 0.2 for LaSOT-ext and TNL2K, and precision, recall, and F-score for VOT-LT2022.

The verifier uses Qwen3-VL-8B-Thinking in a zero-shot setting, while the correction agent uses a fine-tuned Qwen3-VL-4B-Instruct. We use three verification votes with a stride of 4 and three correction rounds. SAM3 remains frozen throughout all experiments. Trigger settings are selected on a held-out SA-V training shard, and fixed for evaluation. Further details can be found in the appendix.

\subsection{Quantitative Results}
Our experiments are designed to validate three questions central to the proposed framework. First, can the correction agent recover targets in regimes where recursive propagation fails? Second, does explicit failure-aware reasoning improve a strong SAM3 propagator on challenging long-video benchmarks? Third, do these gains arise from sparse and well-targeted interventions rather than frequent rewrites of tracker state?

\begin{table}[!htbp]
\centering
\small
\setlength{\tabcolsep}{6pt}
\caption{Corrector-only validation on the held-out corrector-SFT split. Each sample is one query frame from the 7-frame prompt window; results report IoU against ground truth for SAM3 and the corrector on the directly measurable subset (1{,}888 samples).}
\label{tab:val_stratified}
\resizebox{0.7\textwidth}{!}{
\begin{tabular}{lccccc}
\toprule
SAM3 IoU regime & Samples & SAM3 IoU & Corrector IoU & $\Delta$ & Wins / Losses \\
\midrule
$<0.1$ (lost target)        & 347 & 0.024 & 0.612 & +0.587 & 270 / 11 \\
$0.1$--$0.4$ (severe drift) & 808 & 0.224 & 0.727 & +0.503 & 688 / 76 \\
$0.4$--$0.7$ (drift)        & 182 & 0.492 & 0.734 & +0.242 & 120 / 16 \\
$\geq 0.7$ (already good)   & 551 & 0.955 & 0.912 & $-0.043$ & 9 / 40 \\
\bottomrule
\end{tabular}
}
\end{table}

We first evaluate the correction agent in isolation to determine whether it can recover targets once SAM3 has already drifted or lost identity. Table~\ref{tab:val_stratified} reports validation performance on the held-out corrector-SFT split, with one query frame sampled from each 7-frame prompt window. For each query frame, we compare the IoU of SAM3's prediction and the corrector's output against ground truth.
The results show a clear asymmetric pattern. The corrector substantially improves localization once SAM3 has drifted or lost the target entirely. In contrast, when SAM3 is already correct, the corrector provides little benefit and can slightly reduce accuracy. This validates our core motivation: recovery is valuable precisely in the regimes where recursive propagation becomes unreliable, while unnecessary intervention must be avoided through explicit verification.


\begin{table*}[!htbp]
\centering
\small
\caption{LVOS-v2 validation comparison against prior methods. $\mathcal{J}\&\mathcal{F}$ is the overall score; $J_s/F_s$ and $J_u/F_u$ denote seen- and unseen-object region/boundary scores.}
\resizebox{0.5\textwidth}{!}{%
\begin{tabular}{lccccc}
\toprule
Method & $\mathcal{J}\&\mathcal{F}$ & $J_s$ & $F_s$ & $J_u$ & $F_u$ \\
\midrule
LWL & 60.6 & 58.0 & 64.3 & 57.2 & 62.9 \\
RDE & 62.2 & 56.7 & 64.1 & 60.8 & 67.2 \\
STCN & 60.6 & 57.2 & 64.0 & 57.5 & 63.8 \\
XMem & 64.5 & 62.6 & 69.1 & 60.6 & 65.6 \\
SAM 2.1 & 84.1 & 80.7 & 87.4 & 80.6 & 87.7 \\
SAM2.1Long & 85.9 & 81.7 & 88.6 & 83.0 & 90.5 \\
SeC & 86.5 & --- & --- & --- & --- \\
SAM3 & 88.2 & \textbf{85.4} & \textbf{92.9} & 83.3 & 91.5 \\
Dual-Agent (ours) & \textbf{89.4} & 84.9 & 92.7 & \textbf{85.8} & \textbf{94.1} \\
\bottomrule
\end{tabular}
}
\label{tab:lvos_compare}
\end{table*}

\begin{table*}[!htbp]
\centering
\small
\caption{Mechanism-level analysis on SecVOS: (a) direct attribution by SAM3 state at commit time, and (b) controller activity. In (b), \texttt{Verifier}, \texttt{Corrector}, and \texttt{Commit} count verifier calls, corrector calls, and committed corrections; \texttt{Outcome} reports better / worse / neutral changes relative to SAM3.}
\vspace{3pt}
\begin{tabular}{@{}c@{\hspace{1.2em}}c@{}}
\begin{minipage}[t]{0.43\textwidth}
\centering
\footnotesize
\setlength{\tabcolsep}{2pt}
\textbf{(a) Direct Attribution}\par\vspace{2pt}
\begin{tabular}{lccc}
\toprule
State at commit & Count & SAM3 IoU & Dual IoU \\
\midrule
SAM3 missed & 39 & 0.000 & 0.856 \\
SAM3 drifted & 32 & 0.035 & 0.903 \\
SAM3 was correct & 17 & 0.961 & 0.026 \\
\bottomrule
\end{tabular}
\end{minipage}
&
\begin{minipage}[t]{0.46\textwidth}
\centering
\footnotesize
\setlength{\tabcolsep}{2pt}
\textbf{(b) Controller Statistics}\par\vspace{2pt}
\begin{tabular}{lcccc}
\toprule
Trigger & Verifier & Corrector & Commit & Outcome \\
\midrule
Overall & 1360 & 240 & 88 & 67/19/2 \\
\midrule
Eff IoU var & 713 & 120 & 53 & 42/9/2 \\
Occlusion & 508 & 105 & 26 & 19/7/0 \\
Reappear & 139 & 15 & 9 & 6/3/0 \\
\bottomrule
\end{tabular}
\end{minipage}
\end{tabular}
\label{tab:secvos_analysis}
\end{table*}

We then evaluate full closed-loop system on LVOS-v2. Table~\ref{tab:lvos_compare} places the proposed framework against prior LVOS-v2 methods and also reports SAM3 baseline. Besides the overall $\mathcal{J}\&\mathcal{F}$ score, the table includes seen-object and unseen-object breakdowns through $J_s/F_s$ and $J_u/F_u$. The proposed framework improves SAM3 from 88.2 to 89.4 $\mathcal{J}\&\mathcal{F}$ while achieving the highest unseen-object performance. These results show that explicit failure-aware reasoning remains beneficial even on top of a strong propagator. The slight drop on seen subset might be because the verifier can occasionally overrule correct SAM3 predictions.

SecVOS provides the clearest test of the paper's central claim. The benchmark is built around scene discontinuity, reappearance, and abrupt semantic change, so it directly probes the regime in which propagation can be semantically wrong. Table~\ref{tab:secvos_breakdown} reports the scene-complexity breakdown, including no-scene-change, single-scene-change, and multi-scene-change subsets. On this benchmark, the Dual-Agent framework improves $\mathcal{J}\&\mathcal{F}$ from 74.21 to 76.94, with gains increasing as scene complexity grows and the largest improvement observed on the multi-scene-change subset.

Table~\ref{tab:secvos_analysis} provides the mechanism-level view behind this benchmark gain. Table~\ref{tab:secvos_analysis}(b) summarizes controller activity: how often the verifier fires, how often the corrector is invoked, how many corrections are ultimately committed, and how these events break down by trigger type. Its `Outcome' column reports all committed corrections as better / worse / neutral relative to SAM3. The controller is sparse: over the full SecVOS run, it issues 1360 verifier calls, 240 corrector calls, and only 88 committed corrections, so the verifier is invoked on just 4.82\% of unique video frames. On the 502 verifier decisions at frames where the ground-truth box is present, the verifier attains 68.7\% accuracy, 76.8\% precision, and 60.0\% recall, indicating a precision-oriented operating point. Among all 88 commits, 67 improve over SAM3, 19 worsen, and 2 are neutral.

\begin{table*}[!htbp]
\centering
\small
\caption{Performance comparison on SecVOS-160 for different scene-complexity subsets.}
\vspace{3pt}
\setlength{\tabcolsep}{4.5pt}
\renewcommand{\arraystretch}{1.08}
\resizebox{0.7\textwidth}{!}{%
\begin{tabular}{lcccccccccc}
\toprule
\textbf{Method} & \multicolumn{3}{c}{\textbf{No Scene Change}} & \multicolumn{3}{c}{\textbf{Single Scene Change}} & \multicolumn{3}{c}{\textbf{Multi Scene Change}} & \textbf{Overall} \\
\cmidrule(lr){2-4} \cmidrule(lr){5-7} \cmidrule(lr){8-10} \cmidrule(lr){11-11}
& \textit{J\&F} & \textit{J} & \textit{F} & \textit{J\&F} & \textit{J} & \textit{F} & \textit{J\&F} & \textit{J} & \textit{F} & \textit{J\&F} \\
\midrule
XMem & 71.9 & 72.0 & 71.8 & 47.0 & 47.9 & 46.2 & 41.9 & 42.4 & 41.4 & 48.4 \\
DEVA & 71.6 & 71.6 & 71.5 & 48.5 & 48.4 & 48.6 & 46.4 & 46.0 & 46.8 & 49.7 \\
Cutie-base & 72.5 & 72.2 & 72.8 & 53.0 & 52.9 & 53.2 & 48.3 & 47.8 & 48.9 & 52.7 \\
SAM2.1 & 79.4 & 79.1 & 79.7 & 58.5 & 58.2 & 58.8 & 52.4 & 52.1 & 52.6 & 58.2 \\
SAMURAI & 81.8 & 81.6 & 81.9 & 60.6 & 60.6 & 60.7 & 59.3 & 58.9 & 59.7 & 62.2 \\
SAM2.1Long & 81.3 & 81.0 & 81.6 & 61.8 & 61.6 & 62.0 & 58.5 & 58.1 & 58.9 & 62.3 \\
SeC & 84.2 & 83.8 & 84.5 & 69.6 & 69.5 & 69.7 & 67.5 & 67.0 & 68.0 & 70.0 \\
SAM3 baseline & 87.09 & 87.0 & 87.2 & 74.62 & 74.7 & 74.6 & 69.27 & 69.1 & 69.4 & 74.21 \\
Dual-Agent (ours) & \textbf{88.11} & \textbf{87.9} & \textbf{88.3} & \textbf{76.22} & \textbf{76.3} & \textbf{76.1} & \textbf{73.39} & \textbf{73.3} & \textbf{73.5} & \textbf{76.94} \\
\bottomrule
\end{tabular}
}
\label{tab:secvos_breakdown}
\end{table*}
\begin{table*}[!htbp]
\centering
\small
\caption{SecVOS under the Dual-Agent recipe: (a) trigger design, (b) voting, (c) refinement rounds.}
\vspace{3pt}
\begin{minipage}[t]{0.38\textwidth}
\centering
\textbf{(a) Trigger Ablation}\par\vspace{2pt}
\footnotesize
\setlength{\tabcolsep}{2pt}
\begin{tabular}{lccc}
\toprule
Trigger setting & J\&F & J & F \\
\midrule
Occlusion                          & 75.4 & 75.3 & 75.5 \\
+ reappearance                     & 75.6 & 75.5 & 75.7 \\
+ effective-IoU variance           & \textbf{76.94} & \textbf{76.9} & \textbf{77.0} \\
\bottomrule
\end{tabular}
\end{minipage}\hfill
\begin{minipage}[t]{0.38\textwidth}
\centering
\textbf{(b) Voting Ablation}\par\vspace{2pt}
\footnotesize
\setlength{\tabcolsep}{2pt}
\begin{tabular}{ccccc}
\toprule
Vote window & Vote stride & J\&F & J & F \\
\midrule
3 & 4 & \textbf{76.94} & \textbf{76.9} & \textbf{77.0} \\
1 & 4 & 76.8 & 76.9 & 76.7 \\
3 & 2 & 76.4 & 76.4 & 76.4 \\
\bottomrule
\end{tabular}
\end{minipage}\hfill
\begin{minipage}[t]{0.22\textwidth}
\centering
\textbf{(c) Refinement Rounds}\par\vspace{2pt}
\footnotesize
\setlength{\tabcolsep}{2pt}
\begin{tabular}{cccc}
\toprule
Rounds & J\&F & J & F \\
\midrule
3 & \textbf{76.94} & \textbf{76.9} & \textbf{77.0} \\
1 & 75.6 & 75.6 & 75.7 \\
\bottomrule
\end{tabular}
\end{minipage}
\label{tab:secvos_ablation_pair}
\end{table*}

Table~\ref{tab:secvos_analysis}(a) complements this view by attributing committed interventions to the state of SAM3 at the moment of correction. The dominant positive regime is \emph{SAM3 missed}, where SAM3 has effectively lost the object and the commit restores a high-IoU state. The dominant negative regime is \emph{SAM3 was correct}, where an unnecessary intervention overrides a correct track. Taken together, the two parts of Table~\ref{tab:secvos_analysis} show that the improvement comes from selectively rescuing genuine failures rather than frequently rewriting the tracker state.

Table~\ref{tab:secvos_ablation_pair} identifies which parts of the controller are responsible for the SecVOS gain. Table~\ref{tab:secvos_ablation_pair}(a) is a trigger ablation: it starts from occlusion alone, then adds reappearance, and finally adds effective-IoU variance. The largest jump comes from effective-IoU variance, confirming that explicit drift sensing is the main driver of recovery volume. Table~\ref{tab:secvos_ablation_pair}(c) is a refinement-depth ablation and shows that reducing correction from three rounds to one causes a comparable drop, so detecting failure is not enough unless the corrector is allowed to resolve it fully. By contrast, Table~\ref{tab:secvos_ablation_pair}(b) shows that the voting design is comparatively stable: reducing the vote window from three to one changes little, while shrinking the vote stride is mildly harmful. On SecVOS, the load-bearing components are therefore triggered by design and sufficient correction depth, while voting mainly stabilizes the decision process. Overall, the framework improves SecVOS by +2.7 J\&F while intervening on only 88 frames, showing that sparse, targeted recovery is sufficient to substantially improve long-horizon tracking.

\subsection{Further Analysis}

\paragraph{Computational cost.}
Table~\ref{tab:cost_main} quantifies the cost of failure-aware reasoning on a single H100. The full system requires 71.8 seconds per video, compared with 12.2 seconds for SAM3. FP8 verification reduces this to 56.2 seconds while maintaining nearly identical performance (76.80 vs.\ 76.94 $\mathcal{J}\&\mathcal{F}$). Reducing the occlusion polling frequency to once every 24 frames further lowers runtime to 47.7 seconds while retaining 76.41 $\mathcal{J}\&\mathcal{F}$. Thus, verification frequency and precision provide a trade-off between cost and tracking accuracy. The current implementation, however, is not real-time.

\paragraph{Transfer beyond SAM3.}
We next examine whether the proposed framework only improves SAM3.
As shown in Table~\ref{tab:transfer}, Dual-Agent consistently improves every evaluated propagator without retraining, indicating that the proposed failure-recovery mechanism is not tied to a specific tracking architecture. Larger gains are observed for weaker propagators, which present more opportunities for recovery, while SeC and SAM3 also improve despite their stronger propagation performance. The improvements on SeC, which already incorporates VLM-based semantic reasoning, suggest that the verify-and-correct framework complements rather than duplicates semantic propagation, while the gains on Cutie, which exposes no confidence score and thus uses only the occlusion-polling trigger, show that the learned recovery policy is not specific to SAM-based trackers. 

\paragraph{Single-object tracking.}
Dual-Agent consistently improves SAM3 across all three benchmarks. On VOT-LT2022, higher recall yields an F-score improvement, indicating improved recovery after target disappearance. Gains remain consistent on LaSOT-ext and are largest on TNL2K, whose severe appearance changes and long occlusions more frequently require recovery. These results show that the proposed verify-and-correct framework generalizes beyond VOS to standard long-term tracking.

\begin{table}[t]
\caption{Dual-Agent on other frozen propagators, SecVOS-160 ($\mathcal{J}\&\mathcal{F}$). Nothing is retrained.}
\label{tab:transfer}
\centering
\small
\setlength{\tabcolsep}{4pt}
\resizebox{0.8\textwidth}{!}{%
\begin{tabular}{l cc c cc c cc c cc}
\toprule
& \multicolumn{2}{c}{Overall} & & \multicolumn{2}{c}{No scene change} & & \multicolumn{2}{c}{Single change} & & \multicolumn{2}{c}{Multiple changes} \\
\cmidrule{2-3}\cmidrule{5-6}\cmidrule{8-9}\cmidrule{11-12}
Propagator & Base & +Dual-Agent & & Base & +Dual-Agent & & Base & +Dual-Agent & & Base & +Dual-Agent \\
\midrule
Cutie~\citep{cheng2024putting}  & 52.7 & \textbf{59.2} & & 72.5 & \textbf{75.9} & & 53.0 & \textbf{59.5} & & 48.3 & \textbf{56.6} \\
SAM2.1~\citep{ravi2024sam}      & 58.2 & \textbf{65.1} & & 79.4 & \textbf{83.4} & & 58.5 & \textbf{61.6} & & 52.4 & \textbf{61.3} \\
SeC~\citep{zhang2025sec}        & 70.0 & \textbf{72.0} & & 84.2 & \textbf{84.6} & & 69.6 & \textbf{72.2} & & 67.5 & \textbf{70.5} \\
SAM3~\citep{carion2025sam}      & 74.21 & \textbf{76.94} & & 87.09 & \textbf{88.11} & & 74.62 & \textbf{76.22} & & 69.27 & \textbf{73.39} \\
\bottomrule
\end{tabular}}

\end{table}

\begin{table}[t]
\centering
\begin{minipage}[t]{0.4\textwidth}
\centering
\caption{Per-video cost on one H100.}
\label{tab:cost_main}
\footnotesize
\setlength{\tabcolsep}{3pt}
\begin{tabular}{@{}lrrr@{}}
\toprule
Configuration & Sec. & PFLOP & $\mathcal J\&\mathcal F$\\
\midrule
SAM3 & 12.2 & 1.38 & 74.21\\
Dual-Agent (bf16) & 71.8 & 2.22 & \textbf{76.94}\\
Dual-Agent (FP8) & 56.2 & 2.22 & 76.80\\
\quad + poll every 24 & 47.7 & 2.01 & 76.41\\
\bottomrule
\end{tabular}
\end{minipage}\hfill
\begin{minipage}[t]{0.57\textwidth}
\centering
\caption{Single-object tracking.}
\label{tab:tracking}\label{tab:lasotext}\label{tab:tnl2k}\label{tab:votlt}
\footnotesize
\setlength{\tabcolsep}{2.2pt}
\begin{tabular}{@{}l ccc ccc ccc@{}}
\toprule
& \multicolumn{3}{c}{LaSOT-ext} & \multicolumn{3}{c}{TNL2K} & \multicolumn{3}{c}{VOT-LT2022} \\
\cmidrule(lr){2-4}\cmidrule(lr){5-7}\cmidrule(l){8-10}
Method & AUC & P & P$_{\rm norm}$ & AUC & P & P$_{\rm norm}$ & Pr & Re & F \\
\midrule
SAM3 & 63.7 & 74.8 & 72.9 & 65.2 & 72.8 & 74.3 & \textbf{63.2} & 57.4 & 60.1 \\
Dual-Agent & \textbf{64.3} & \textbf{75.5} & \textbf{73.5} & \textbf{66.5} & \textbf{74.1} & \textbf{75.7} & 61.0 & \textbf{61.3} & \textbf{61.2} \\
\bottomrule
\end{tabular}
\end{minipage}
\end{table}

\subsection{Qualitative Results}
\begin{figure}[!htbp]
    \centering
    
    \includegraphics[width=0.75\textwidth]{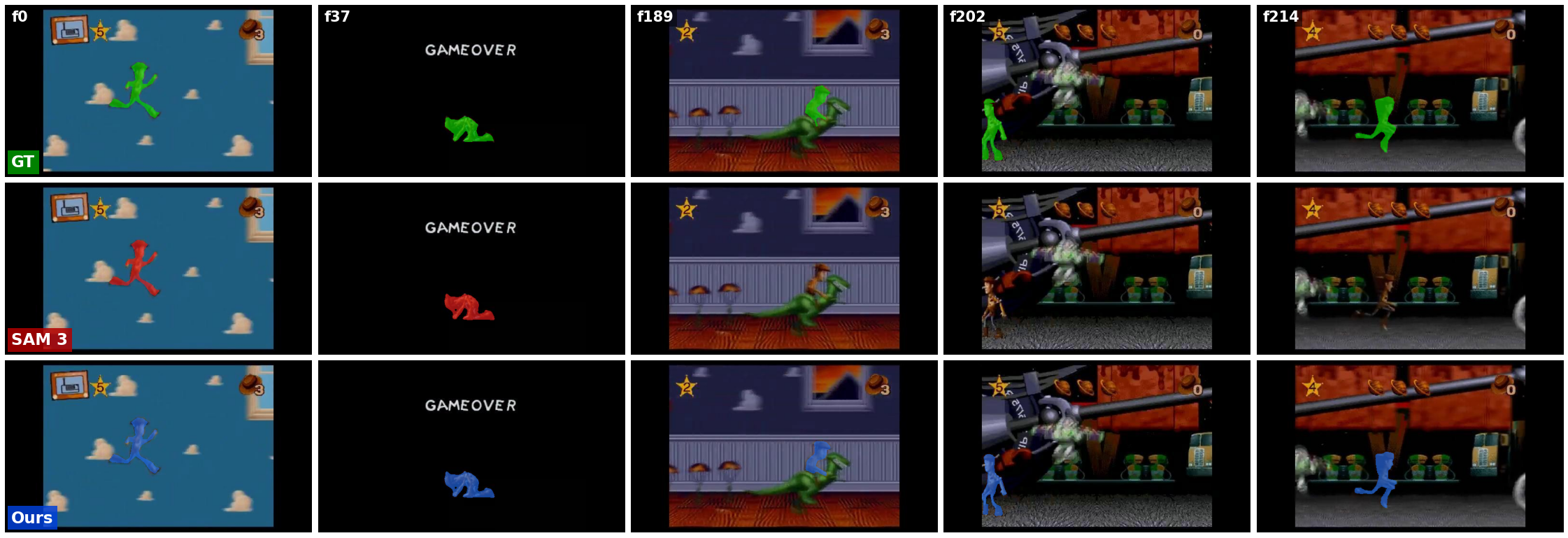}
    
    \vspace{0.5em}
    \hrule
    \vspace{0.5em}
    
    \includegraphics[width=0.75\textwidth]{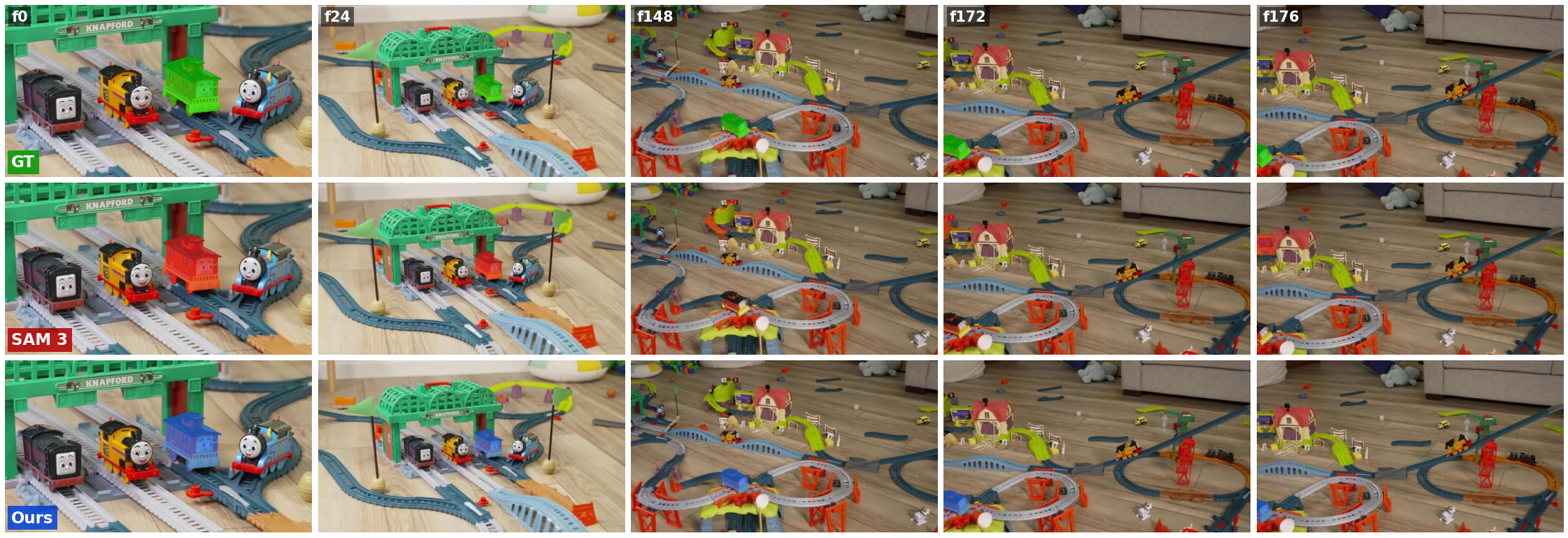}
    
    \caption{Qualitative recovery on two SAM3 failure modes.
\textbf{Top:} reappearance after a long absence;
\textbf{Bottom:} drift on the target toy train in a cluttered scene with similar distractors.
For each example, rows show ground truth (top), the SAM3 baseline
(middle), and our Dual-Agent framework (bottom).}
    \label{fig:qualitative}
\end{figure}

Figure~\ref{fig:qualitative} illustrates two representative failure modes recovered by the proposed Dual-Agent framework. The top sequence shows long-horizon \emph{reappearance}: after the target leaves the scene and re-enters with significant pose and motion change, SAM3 fails to recover it. The verifier detects the propagated prediction as inconsistent with the target, after which the corrector re-localizes the object and restores stable propagation.
The bottom sequence shows \emph{drift in cluttered scenes}. As the target train moves through visually similar distractors, SAM3 latches onto a neighboring object and its propagated state becomes unstable. The effective-IoU variance trigger activates the verifier, which confirms the failure. The corrector then recovers the correct target allowing propagation to continue successfully.

\section{Conclusion}
We present a Dual-Agent framework that uses SAM3 as the propagator and augments it with explicit verification and selective recovery. By separating reliability estimation from target correction, the framework turns a SAM3-style propagator into a failure-aware tracking system. The verifier reasons over short spatio-temporal context to detect unreliable predictions, while a learned correction agent recovers the target only when intervention is warranted. Evaluation across segmentation and tracking benchmarks demonstrates that robust tracking requires not only stronger propagation, but also explicit reasoning about when propagation should no longer be trusted. In highly ambiguous scenarios, incorrect reasoning may lead to missed failures or unnecessary corrections, while the additional verification stage also increases computational overhead and latency.

\FloatBarrier
\begin{ack}
The work described in this paper was conducted in the framework of the Graduate School 2543/1 ``Intraoperative Multi-Sensory Tissue Differentiation in Oncology'' (project ID 40947457) funded by the German Research Foundation (DFG - Deutsche Forschungsgemeinschaft). This work has been supported by the Deutsche Forschungsgemeinschaft (DFG) -- EXC number 2064/1 -- Project number 390727645 and SFB 1233, TP 2, Project number 276693517. The authors thank International Max Planck Research School for Intelligent Systems (IMPRS-IS) for supporting V.~Bundele, M.~Hosseinzadeh and S.~Agrawal.
\end{ack}

{\small
\bibliographystyle{plainnat}
\bibliography{bibliography}
}

\clearpage

\title{Supplementary Material:\\When Trackers Fail: VLM-Guided Verification and Recovery for Robust Video Object Segmentation}

\author{%
  Anonymous Authors
}


\appendix
\section{Implementation Details}
\textit{Verification agent.} We use Qwen3-VL-8B-Thinking in its native thinking mode with no fine-tuning, prompted with a structured tracking template. Each verifier query is a 7-image visual context laid out as 2 ground-truth exemplar frames, 2 uniformly sampled context frames, 2 recent context frames at stride 4, and 1 query frame. The verifier votes 3 times per call with a vote stride of 4 frames; the majority verdict is committed. In our experiments, we set $\tau_{\mathrm{conf}}=0.7$, $K=3$, $M=12$ and $N=8$.

\textit{Correction agent.} We use Qwen3-VL-4B-Instruct adapted with LoRA (rank 16, $\alpha = 32$, dropout 0.05, applied to all linear layers; vision tower frozen, and keep the multimodal projector trainable). The corrector is fine-tuned with AdamW at learning rate $2 \times 10^{-5}$ on a cosine schedule with warmup ratio 0.05 and weight decay 0.01, for 2 epochs at effective batch size 16 (4 GPUs $\times$ batch 2 $\times$ gradient-accumulation 2), with bf16 mixed-precision training, gradient checkpointing, a 16{,}384-token sequence cutoff, and a 401{,}408-pixel image budget matched to inference. We fine-tune the VLM on 4 H100s for around a day. The corrector takes the same 7-image layout as the verifier and is invoked in no-think mode using a visible-plus-bounding-box prompt template.

\textit{Selective triggering.} The closed-loop controller fires the verifier on three signals applied in disjunction: occlusion polling (one query every 12 consecutive occluded frames), reappearance probing (one query when the propagator recovers a target after an absence, with cooldown 8 frames), and a streaming effective-IoU variance trigger (with a 4-frame window, threshold 0.15, persistence 2, and a minimum spacing of 8 frames between consecutive fires). When the verifier returns ``incorrect,'' the corrector is invoked with up to 3 retry rounds.

\textit{Propagator.} SAM3 remains entirely frozen across all experiments; the Dual-Agent framework operates as an external control loop that selectively rewrites SAM3's state.

\section{Dataset Details}
\label{sec:supp_datasets}

This appendix gives the full inventory of every benchmark referenced in the main paper, including sequence and object counts, annotation density, and evaluation protocols. 

\subsection{Benchmark Inventory}

\begin{table}[!htbp]
\centering
\small
\setlength{\tabcolsep}{4pt}
\caption{Benchmarks and splits used in the paper. The annotation column distinguishes segmentation masks from tracking boxes.}
\label{tab:supp_dataset_inventory}
\begin{tabular}{llccl}
\toprule
Benchmark & Split & Sequences & Objects & Frames / annotations \\
\midrule
LVOS-v2     & valid             & 140 & 239 & every frame                 \\
SecVOS      & full              & 160 & 182 & every frame  \\
LaSOT-ext & full & 150 & 150 & per-frame boxes  \\
TNL2K & test & 700 & 700 & per-frame boxes  \\
VOT-LT2022 & full & 50 & 50 & per-frame boxes \\
\bottomrule
\end{tabular}
\end{table}

\paragraph{LVOS-v2.}
We evaluate on the official LVOS-v2 validation split, which consists of 140 long sequences containing 239 annotated objects in total. Annotations are dense: every video frame has a corresponding ground-truth mask. The benchmark is designed to stress long-horizon memory. We report results on the standard evaluation harness (\texttt{lvos-evaluation}), which produces \(\mathcal{J}_s\), \(\mathcal{F}_s\), \(\mathcal{J}_u\), \(\mathcal{F}_u\), and overall \(\mathcal{J}\&\mathcal{F}\) on the seen / unseen splits.

\paragraph{SecVOS.}
SecVOS is a 160-sequence benchmark with 182 annotated objects, designed around multi-shot discontinuity, scene transitions, and target reappearance across semantically distinct contexts. We use the standard evaluation protocol from the SeC paper, which scores J\&F on the full benchmark and additionally stratifies the score into three scene-cut buckets: \emph{no scene change}, \emph{single scene change}, and \emph{multi scene change}.

\paragraph{Single-object tracking benchmarks.}
LaSOT-ext evaluates long-term tracking on 150 sequences, TNL2K on its 700-sequence test split, and VOT-LT2022 on 50 long-term sequences with target disappearance and reappearance. Each sequence contains one target, specified by its first-frame bounding box; TNL2K language descriptions are not used. LaSOT-ext and TNL2K use success AUC, 20-pixel precision, and normalized precision at 0.2. VOT-LT2022 uses the official long-term precision, recall, and F-score protocol. 

\section{Training Data Details}
\label{sec:supp_training}

The correction agent is trained on a SFT corpus that we generate ourselves from SAM3 propagation rollouts on SA-V training videos. 

\subsection{Dataset Identity and Sizes}

The corpus is split sequence-stratified at generation time into a 95\% train shard and a 5\% validation shard, yielding 75{,}341 training samples and 4{,}116 validation samples for a total of 79{,}457 supervised correction examples. The split is deterministic (\texttt{seed=42}) and stratified by source video so that no SA-V sequence appears in both shards: this prevents the corrector from receiving an inflated validation score by memorizing per-video patterns. Because the corpus is sequence-stratified rather than frame-stratified, validation-set samples test temporal generalization within unseen sequences from the same SA-V shards.

\subsection{Per-Sample Schema}

\begin{figure}[!htbp]
    \centering
    
    \includegraphics[width=0.9\textwidth]{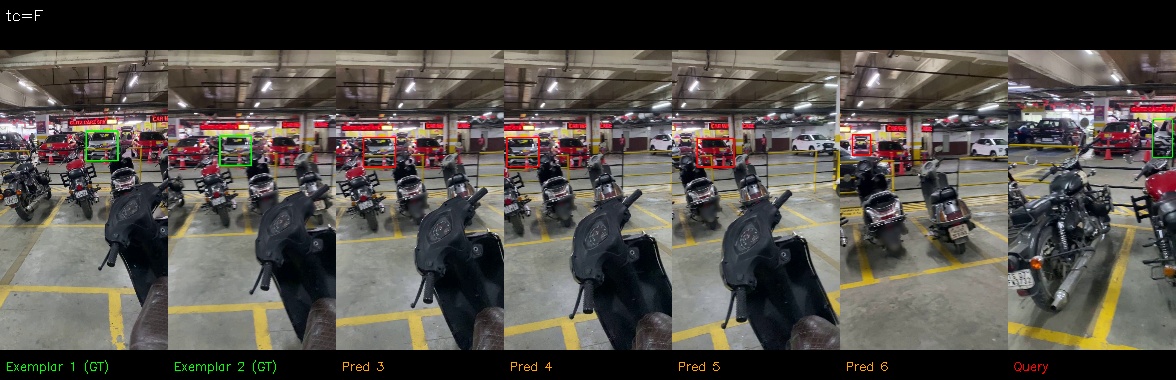}

    \caption{An example from the dataset.}
    \label{fig:supp_example}
\end{figure}

Each supervised example encodes a closed-loop correction decision. As shown in Figure \ref{fig:supp_example}, the visual context is laid out as a fixed seven-frame window:
\begin{itemize}
\item \textbf{2 ground-truth exemplar frames} drawn from the early part of the sequence, where the target identity is unambiguous.
\item \textbf{2 uniform context frames} sampled at uniform stride across the elapsed sequence to capture long-range appearance changes.
\item \textbf{2 recent context frames} immediately preceding the query, capturing short-term motion and occlusion state.
\item \textbf{1 query frame} at which the corrector must predict whether the target is visible and, if so, where.
\end{itemize}

The supervised target at the query frame is a structured JSON object containing a \texttt{visible} boolean and, when visible, a normalized \texttt{box\_2d}. Outputs are produced in a no-think format.

\subsection{Dataset Analysis}
\label{sec:supp_dataset_analysis}

We characterize the corrector-SFT corpus along three axes that together describe what the corrector is supervised to learn: (i)~the distribution of failure modes the corrector must handle, (ii)~the joint distribution of SAM3 correctness and target visibility, and (iii)~the quality of the visual context the corrector reasons over.

\subsubsection{Failure-Mode Composition}

The corrector is trained to recognize and recover from a small set of structurally distinct failure modes that the propagator exhibits at deployment. Table~\ref{tab:supp_sample_types} reports the partition of the corpus by failure type; the eleven categories are mutually exclusive and together account for the full 79{,}457 samples.

More than half of the corpus (56.4\%) targets the override regime: \texttt{false\_negative} (17.8\%) supervises the corrector to localize from context when SAM3 has produced no box at all, \texttt{low\_iou} (15.1\%) supervises refinement when SAM3 has drifted, \texttt{perturbed} (15.5\%) trains the corrector to remain stable when its own context input is jittered, and three identity-switch variants (8.0\% combined) train it to resist confusion from co-occurring distractor objects. The remaining samples balance this with non-override supervision: \texttt{correct} (14.6\%) teaches the corrector to mirror SAM3 when SAM3 is already right, \texttt{both\_none} (13.5\%) teaches graceful behavior when there is nothing to track, and the two negative-label categories (\texttt{false\_positive} 4.4\% and \texttt{synthetic\_fp} 6.0\%) train suppression of spurious detections. The deliberate inclusion of mirror-SAM3 supervision is what prevents the corrector from collapsing into an ``always override'' degenerate solution at deployment time, where it would pollute the propagator's memory with redundant rewrites of already-correct frames.

\begin{table}[!htbp]
\centering
\small
\setlength{\tabcolsep}{6pt}
\caption{Sample-type distribution of the corrector-SFT corpus (79{,}457 samples).}
\label{tab:supp_sample_types}
\setlength{\tabcolsep}{4pt}
\begin{tabular}{lrrl}
\toprule
Sample type & Count & Share & Supervised behavior \\
\midrule
\texttt{false\_negative}             & 14{,}131 & 17.8\% & emit a fresh visible-true box from context \\
\texttt{perturbed}                   & 12{,}349 & 15.5\% & recover from jittered context boxes \\
\texttt{low\_iou}                    & 12{,}011 & 15.1\% & emit a refined box over a drifted SAM3 box \\
\texttt{correct}                     & 11{,}618 & 14.6\% & mirror SAM3's already-correct prediction \\
\texttt{both\_none}                  & 10{,}746 & 13.5\% & emit visible-false on absent target \\
\texttt{synthetic\_fp}               &  4{,}807 &  6.0\% & reject artificially-injected false positive \\
\texttt{paired\_occluded}            &  4{,}000 &  5.0\% & decision on occlusion-onset pairs \\
\texttt{false\_positive}             &  3{,}478 &  4.4\% & emit visible-false to suppress SAM3 hallucination \\
\texttt{id\_switch (false\_negative)} & 2{,}926 &  3.7\% & resist distractor when SAM3 missed the target \\
\texttt{id\_switch (low\_iou)}        & 2{,}753 &  3.5\% & resist distractor when SAM3 drifted \\
\texttt{id\_switch (false\_positive)} &    638 &  0.8\% & resist distractor when SAM3 hallucinated \\
\midrule
Total                                 & 79{,}457 & 100\% & \\
\bottomrule
\end{tabular}
\end{table}

\subsubsection{Joint Distribution of SAM3 Correctness and Visibility}

A coarser but more diagnostic view of the corpus is the cross-tabulation of the supervised label (whether SAM3 was correct at the query frame) against the query-frame visibility (whether the target is actually there). Table~\ref{tab:supp_class_vis} reports this two-by-two split.

Two observations follow from this table. First, the marginal class balance --- 33.5\% correct versus 66.5\% incorrect --- indicates that the corrector sees twice as many failure cases as it sees easy mirror cases, which is the right proportion for a corrector whose job is to repair propagation rather than describe it. Second, the cross-tabulation reveals a subtle but deliberate skew: when the target is visible, SAM3 is correct only 22.2\% of the time, but when the target is absent, SAM3 is correct 60.1\% of the time. In other words, the corpus oversamples the regime where SAM3 has produced a confident-looking box on a frame whose target is missing or substantially wrong, which is exactly the regime where override has the largest expected payoff. The smaller correct-on-absent slice (17.9\% of the corpus) is what teaches the corrector to recognize correct-by-default behavior and avoid spurious overrides on absent targets.

\begin{table}[!htbp]
\centering
\small
\setlength{\tabcolsep}{8pt}
\caption{Class balance (SAM3 correct / incorrect) cross-tabulated against query-frame visibility, on all 79{,}457 samples. Rows are the supervised label; columns are the visibility of target at query frame.}
\label{tab:supp_class_vis}
\begin{tabular}{lrrr}
\toprule
SAM3 was \ldots & Visible target & Absent target & Total \\
\midrule
Correct          & 12{,}385 (15.6\%) & 14{,}223 (17.9\%) & 26{,}608 (33.5\%) \\
Incorrect        & 43{,}403 (54.6\%) &  9{,}446 (11.9\%) & 52{,}849 (66.5\%) \\
\midrule
Total            & 55{,}788 (70.2\%) & 23{,}669 (29.8\%) & 79{,}457 (100\%) \\
\bottomrule
\end{tabular}
\end{table}

\subsubsection{Quality of the Visual Context}

The corrector reasons over a fixed seven-frame visual context (two early-sequence ground-truth exemplars, two uniformly sampled context frames, two recent context frames, and one query frame), and the quality of this context directly determines what reasoning the corrector can perform. We characterize context quality at three granularities: per frame, per sample, and per failure type.

Table~\ref{tab:supp_ctx_iou_hist} reports the per-frame IoU histogram across all 249{,}905 non-exemplar context frames in the corpus. The distribution is strongly bimodal: 43.0\% of context frames are in the high-quality bucket $[0.95,~1.01)$ where SAM3 has accurately tracked the target, and 18.2\% are in the lowest bucket $[0.0,~0.10)$ where SAM3 has effectively lost it. Only 16.5\% of frames fall into the wide middle band $[0.30,~0.85)$, indicating that frames are either clearly clean or clearly degraded. 

Aggregating per-frame IoU into a per-sample difficulty label (Table~\ref{tab:supp_difficulty}) gives a complementary view of how challenging each training example is. Almost two-thirds of samples (64.1\%) are \emph{easy} in the sense that the corrector has at least two clean context frames to anchor on; one in six (16.8\%) is \emph{hard} with only a single good context frame; 3.8\% are \emph{noisy} with no good context; and 15.3\% have \emph{no context frames at all}, forcing the corrector to rely entirely on the two exemplar frames. 

Stratifying mean context IoU by failure type (Table~\ref{tab:supp_ctx_iou_by_type}) reveals where context is hardest. The mean drops monotonically from 0.924 on \texttt{correct} samples (where SAM3 is right and context is clean) to 0.272 on \texttt{id\_switch} samples, where the seven-frame window is constructed to include a confounding object. Because identity-switch supervision is delivered with deliberately noisy context, the corrector cannot rely on a single ``trust the recent box'' heuristic for these cases --- it must use the early-sequence exemplars to anchor identity. The \texttt{low\_iou} (0.536) and \texttt{false\_positive} (0.495) categories sit in the middle, supervising the corrector to override SAM3 even when context tokens themselves provide a noisy version of the failed prediction.

\begin{table}[!htbp]
\centering
\small
\setlength{\tabcolsep}{8pt}
\caption{Histogram of context-frame SAM3 IoU against ground truth across all 249{,}905 non-exemplar context frames in the corpus. Mean IoU 0.683, median 0.925, standard deviation 0.388.}
\label{tab:supp_ctx_iou_hist}
\begin{tabular}{lrr}
\toprule
IoU bucket & Frames & Share \\
\midrule
$[0.00,~0.10)$ & 45{,}598 & 18.2\% \\
$[0.10,~0.30)$ & 15{,}763 &  6.3\% \\
$[0.30,~0.50)$ & 13{,}595 &  5.4\% \\
$[0.50,~0.70)$ & 11{,}646 &  4.7\% \\
$[0.70,~0.85)$ & 15{,}922 &  6.4\% \\
$[0.85,~0.95)$ & 39{,}947 & 16.0\% \\
$[0.95,~1.01)$ & 107{,}434 & 43.0\% \\
\bottomrule
\end{tabular}
\end{table}

\begin{table}[!htbp]
\centering
\small
\setlength{\tabcolsep}{8pt}
\caption{Per-sample difficulty distribution, classified by how many usable frames the sample has.}
\label{tab:supp_difficulty}
\begin{tabular}{lrrl}
\toprule
Difficulty & Samples & Share & Definition \\
\midrule
easy        & 50{,}948 & 64.1\% & at least 2 context frames with IoU $> 0.5$ \\
hard        & 13{,}357 & 16.8\% & one good context frame \\
noisy       &  3{,}000 &  3.8\% & no good context frame \\
no\_context & 12{,}152 & 15.3\% & no context frames \\
\bottomrule
\end{tabular}
\end{table}

\begin{table}[!htbp]
\centering
\small
\setlength{\tabcolsep}{8pt}
\caption{Mean context-frame SAM3-vs-ground-truth IoU stratified by sample type (excluding the exemplar frames). Lower mean IoU corresponds to noisier context the corrector must reason over.}
\label{tab:supp_ctx_iou_by_type}
\begin{tabular}{lrrr}
\toprule
Sample type & N & Mean IoU & Median IoU \\
\midrule
\texttt{correct}          &  9{,}531 & 0.924 & 0.966 \\
\texttt{both\_none}       &  9{,}723 & 0.784 & 0.888 \\
\texttt{perturbed}        &  9{,}736 & 0.774 & 0.895 \\
\texttt{synthetic\_fp}    &  4{,}257 & 0.762 & 0.873 \\
\texttt{false\_negative}  &  9{,}989 & 0.749 & 0.844 \\
\texttt{paired\_occluded} &  3{,}365 & 0.724 & 0.811 \\
\texttt{low\_iou}         & 10{,}996 & 0.536 & 0.511 \\
\texttt{false\_positive}  &  3{,}391 & 0.495 & 0.458 \\
\texttt{id\_switch}       &  6{,}317 & 0.272 & 0.316 \\
\bottomrule
\end{tabular}
\end{table}
\FloatBarrier
\newpage
\section{Verification Agent Prompt}
Our verification agent is a Qwen3-VL-8B-Thinking model, which receives the following prompt:
\begin{lstlisting}[basicstyle=\ttfamily\tiny, breaklines=true]

You are assessing a single-object tracker. You see VIDEO FRAMES with overlays:
- First exemplar frame(s): GREEN box marks the TARGET object's true appearance.
- Context frames: RED box = tracker prediction (may be wrong); no RED box = tracker says target absent.
- Query frame (last): RED box (or none) = the prediction you must assess.

Describe the REAL OBJECTS inside the boxes, never the box colour itself. Reason from what you SEE in the boxes.

NOTES ON THE FRAME LIST:
- (t=N) is the source video-frame index. The same (t=N) may legitimately appear twice if an exemplar coincides with a context sample — treat them as the SAME physical frame and move on.
- "earlier context" frames are far in the past, "recent context" frames are just before the query.

YOUR TASK on the QUERY frame is the discrimination question:
  "Does the tracker correctly track the GREEN target instance in this query frame?"

Answer it by following these steps. Do NOT skip step (a):

(a) LOCATE THE TARGET IN THE QUERY IMAGE FIRST.
    Before looking at the red box at all, scan the query image and find where the
    target object (the one in the GREEN exemplar) actually is. Possible outcomes:
      (a.1) Target is fully visible somewhere in the query.
      (a.2) Target is partially visible — some of it is out-of-frame, occluded by a
            person/object/the frame edge, or behind something. Note WHICH PART is
            visible (head only? legs only? a thin sliver at the edge?).
      (a.3) Target is genuinely absent — fully occluded, fully out-of-frame, gone.
            ONLY pick this if you cannot find the target anywhere in the image.

(b) NOW LOOK AT THE TRACKER OUTPUT.
    Two cases:
      Case I — A RED box IS shown on the query:
        * If (a.1) or (a.2): does the red box overlap the target's currently-visible
          extent by at least HALF? Tight head-only / legs-only / thin-sliver crops
          are CORRECT *if and only if* they sit on the visible part of the target.
          A red box on empty floor / sky / wall / a different object (e.g. a balloon,
          a shoe, another person, the ground) when the target is visible elsewhere
          → INCORRECT (this is an ID switch / drift / false positive even though
          the box exists). State explicitly what is inside the box.
        * If (a.3): the tracker drew a box when the target is absent → INCORRECT
          (false positive).
      Case II — NO RED box on the query (the prompt says "no tracker prediction
      shown"):
        * If (a.1) or (a.2): the target IS visible but the tracker abstained →
          INCORRECT (missed visible target).
        * If (a.3): correct abstention → CORRECT.

(c) VERDICT.
    Be DECISIVE. Default to CORRECT only when you genuinely cannot tell.
    Flag INCORRECT whenever step (b) produced a clear, named failure — even if it
    feels harsh. The downstream corrector handles the actual fix.

EDGE-CASE NOTES:
- Edge-of-frame thin RED boxes (touching x=0 or y=0 or a frame edge) usually mean
  the target is exiting the frame. If you saw the target moving toward that edge
  in the recent context, the tiny box at the edge is almost certainly STILL ON
  the partially-visible target → CORRECT. Do not dismiss it as "on empty space"
  unless you can confirm the target is not visible at that edge at all.
- A RED box that sits on a clearly different object (different colour, different
  category, distractor that crossed paths with the target, the ground) while
  the target is visible elsewhere in the frame → INCORRECT (ID switch / drift).
- Imperfect tightness on a fully-visible target (box covers ~70-90% of the
  target) → CORRECT. The corrector cares about IoU>=0.5, not pixel-perfect.

OUTPUT FORMAT (MANDATORY — your response is parsed automatically):
Your final line MUST be of the form:
<answer>{"tracking_correct": true}</answer>
or
<answer>{"tracking_correct": false}</answer>
Do NOT produce any other JSON keys. Do NOT continue reasoning after </answer>.

LENGTH BUDGET: keep your `<think>` block under ~250 words. If you find yourself
rephrasing the same observation, STOP and emit your verdict.

Decision rules summary:
- TRUE: red box covers ≥50% of the target's visible extent, OR no box AND target genuinely absent.
- FALSE: target visible but no box; box on a different object; box on empty space when target is visible elsewhere; box covering well under half of a fully-visible target.

Case 1 — Tracking is correct:
<think>your reasoning here</think>
<answer>{"tracking_correct": true}</answer>

Case 2 — Tracking is incorrect:
<think>your reasoning here</think>
<answer>{"tracking_correct": false}</answer>
\end{lstlisting}
\section{Further Qualitative Results}
\begin{figure}[!htbp]
    \centering
    
    \includegraphics[width=0.9\textwidth]{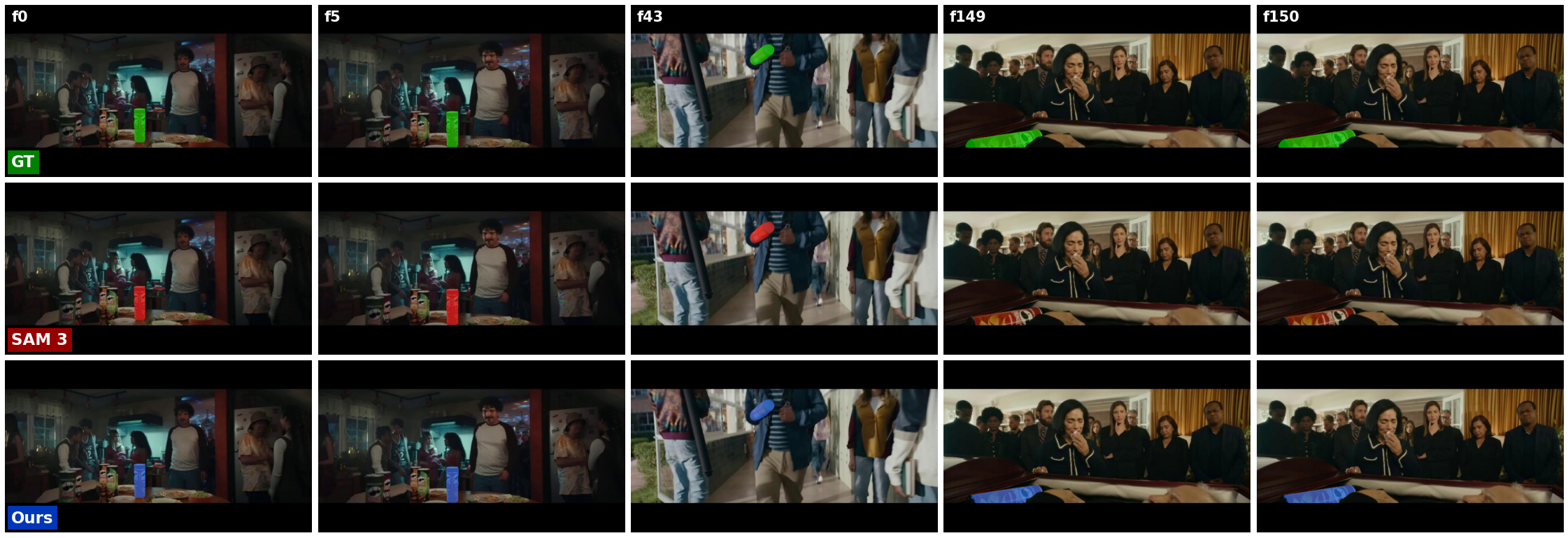}
    
    \vspace{0.5em}
    \hrule
    \vspace{0.5em}
    
    \includegraphics[width=0.9\textwidth]{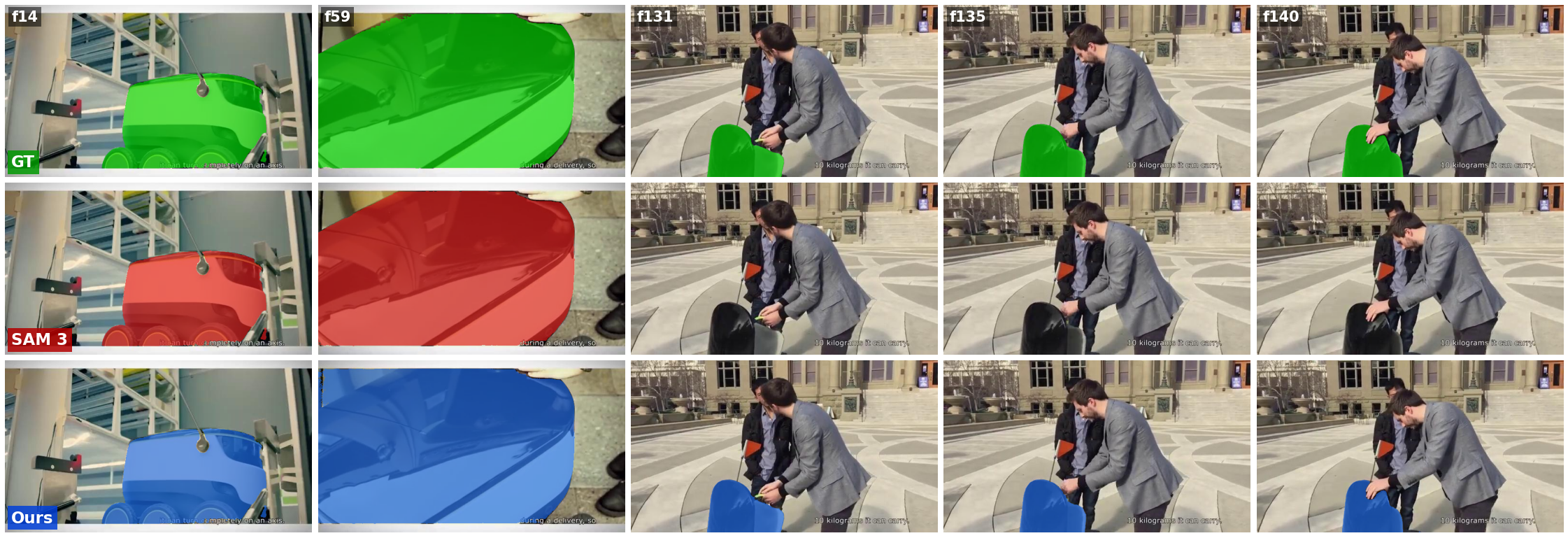}

    \vspace{0.5em}
    \hrule
    \vspace{0.5em}
    
    \includegraphics[width=0.9\textwidth]{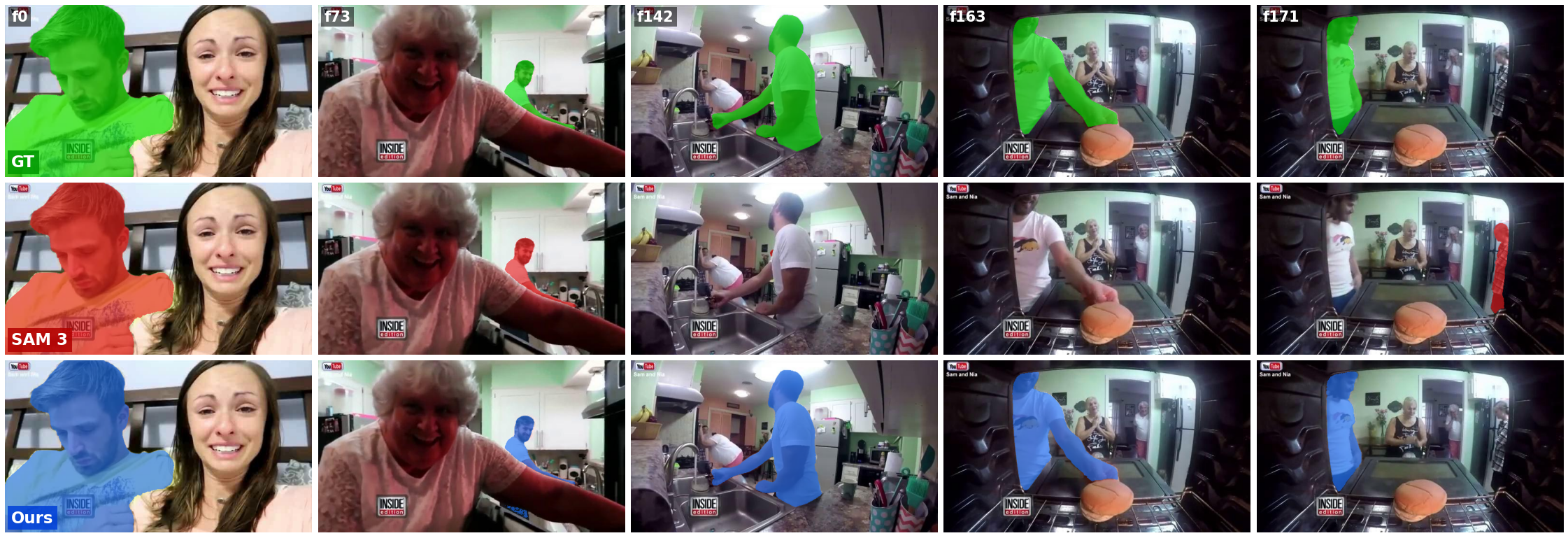}
    
    \caption{Further qualitative examples of our model vs. SAM3.
\textbf{Top:} As the target object partially reappears after a long time in a largely different context, SAM3's score for the object drops, and it fails to identify it; this triggers our recovery mechanism, and the object is re-identified by our dual agents.
\textbf{Middle:} Upon partial appearance of the object, SAM3 fails to find the target object, due to its significant shift from its prior appearance. However, it is recovered by a further VLM-based retrieval through our designed agents. \textbf{Bottom:} Due to the change of the view of the scene, a different pose of the person appears in the video, resulting in the object not being found by SAM3. Our verification agent is thus triggered and detects the need for correction, with the recovery applied afterwards by the second correction agent.
For each example, rows show ground truth (top), the SAM3 baseline
(middle), and our Dual-Agent framework (bottom).}
    \label{fig:supp_qualitative}
\end{figure}

More visual tracking outputs are illustrated in Figure \ref{fig:supp_qualitative}.

\clearpage

\section{Verification and Commit-Gate Ablation}
\label{app:agent_gate}

\begin{table}[h]
\caption{Verification and commitment ($\mathcal{J}\&\mathcal{F}$). (a) Bypassing verification returns SecVOS-160 to the frozen-propagator level. (b) Removing the commit gate keeps most of the gain on both benchmarks.}
\label{tab:reason_vs_gate}
\begin{center}
\small
\resizebox{\textwidth}{!}{%
\begin{tabular}{@{}c@{\hspace{2em}}c@{}}
\begin{tabular}[t]{@{}c@{}}
\textbf{(a) Agent ablation}\\[2pt]
\begin{tabular}{lc}
\toprule
Configuration & SecVOS-160 \\
\midrule
SAM3 & 74.21 \\
Corrector at every trigger (verifier skipped) & 74.10 \\
Dual-Agent (ours) & \textbf{76.94} \\
\bottomrule
\end{tabular}
\end{tabular}
&
\begin{tabular}[t]{@{}c@{}}
\textbf{(b) Commit-gate ablation}\\[2pt]
\begin{tabular}{lccc}
\toprule
Configuration & Gate & SecVOS-160 & LVOS-v2 \\
\midrule
SAM3 & -- & 74.21 & 88.2 \\
Dual-Agent & off & 75.90 & 89.3 \\
Dual-Agent (ours) & on & \textbf{76.94} & \textbf{89.4} \\
\bottomrule
\end{tabular}
\end{tabular}
\end{tabular}}
\end{center}
\end{table}

\paragraph{Reasoning vs commit gating.}
Table~\ref{tab:reason_vs_gate} disentangles the roles of verifier and commit gating. Without verifier, performance on SecVOS drops from 76.94 to 74.10 $\mathcal{J}\&\mathcal{F}$, essentially returning to the SAM3 baseline (74.21), despite retaining the commit gate. In contrast, removing the commit gate retains most of the improvement. This indicates that the gains cannot be explained by conservative acceptance alone: identifying when the tracker has failed and recovery is needed is a key component of the framework. We further analyze correction uncertainty and harmful interventions in Appendix~\ref{app:recovery_reliability}.

\section{Correction Reliability and Harmful Interventions}
\label{app:recovery_reliability}
\paragraph{Seen and unseen LVOS objects.}
An audit against matched SAM3-only trajectories covers all 239 objects in 140 validation videos. Table~\ref{tab:lvos_interventions} separates untouched objects from those modified by the controller. Eight of 202 seen objects regress, a harmful-intervention rate of 4.0\% over that split. The harmful cases concentrate in crowded scenes with multiple similar instances. Among the six modified unseen objects, five improve. Thus, the split disparity is associated with a small, different distribution of intervention outcomes, rather than a broad degradation of seen-object tracking.

\begin{table}[htbp]\centering\small
\caption{Object-level intervention outcomes on LVOS-v2. Improved/unchanged/regressed counts are among modified objects, not counts of individual commits.}
\label{tab:lvos_interventions}
\begin{tabular}{lrrrrr}\toprule
Split & Objects & Untouched & Improved & Unchanged & Regressed\\\midrule
Seen & 202 & 185 & 7 & 2 & 8\\
Unseen & 37 & 31 & 5 & 0 & 1\\\bottomrule
\end{tabular}\end{table}

\begin{figure}[htbp]
\centering
\includegraphics[width=\textwidth]{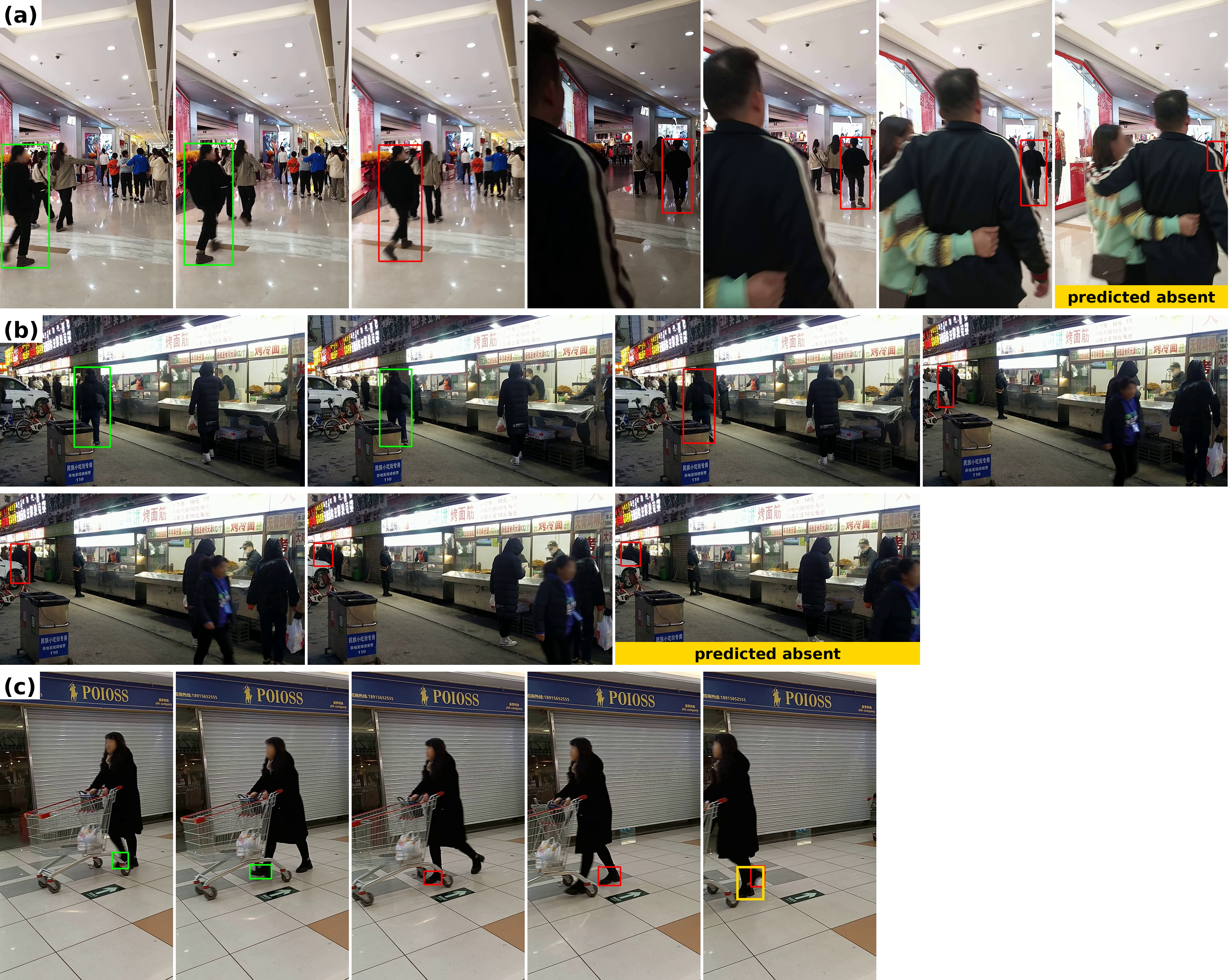}
\caption{Representative failures of the unconditional corrector on already-accurate SAM3 predictions (Table~\ref{tab:val_stratified}, IoU $\geq 0.7$). Each row shows the full corrector input: ground-truth exemplars (green), SAM3 context (red), and the query, with the corrector's output in yellow. (a) A target leaving the frame edge and (b) a small distant target are wrongly predicted absent. (c) The box drifts to a look-alike (the other shoe).}
\label{fig:already_good_failures}
\end{figure}

\paragraph{Failure modes on already-accurate predictions.}
The already-good cell in Table~\ref{tab:val_stratified} is a corrector-only stress test: recovery is applied unconditionally to 551 samples with mean SAM3 IoU 0.955. It does not use the deployed verifier or selective commit policy. The median IoU change is 0.00, and 91.1\% of examples remain within $\pm0.05$. The negative mean ($-0.043$) is driven mainly by 26 failures---17 incorrect absence predictions and nine distractor captures---accounting for 86.6\% of the total loss. Figure~\ref{fig:already_good_failures} shows representative cases. It motivates selective intervention while also showing that acceptance gates do not guarantee harmless correction.

\FloatBarrier
\section{Scope and Limitations}
\label{app:rebuttal_limits}
The current implementation is not real-time on a single GPU. Average runtime does not characterize worst-case intervention latency. Incorrect verification or correction can override an accurate track and affect subsequent memory, as quantified in Appendix~\ref{app:recovery_reliability}; the gates reduce opportunities for harm but provide no formal safety guarantee.

Transfer experiments cover several propagators and natural-video benchmarks. They do not establish performance in specialized domains such as endoscopic imagery, where additional failure mining or model adaptation may be needed. We cannot verify video-level overlap between the foundation models' pretraining data and the evaluation benchmarks. 

\clearpage
\section*{NeurIPS Paper Checklist}

The checklist is designed to encourage best practices for responsible machine learning research, addressing issues of reproducibility, transparency, research ethics, and societal impact. Do not remove the checklist: {\bf The papers not including the checklist will be desk rejected.} The checklist should follow the references and follow the (optional) supplemental material.  The checklist does NOT count towards the page
limit. 

Please read the checklist guidelines carefully for information on how to answer these questions. For each question in the checklist:
\begin{itemize}
    \item You should answer \answerYes{}, \answerNo{}, or \answerNA{}.
    \item \answerNA{} means either that the question is Not Applicable for that particular paper or the relevant information is Not Available.
    \item Please provide a short (1--2 sentence) justification right after your answer (even for \answerNA). 
\end{itemize}

{\bf The checklist answers are an integral part of your paper submission.} They are visible to the reviewers, area chairs, senior area chairs, and ethics reviewers. You will also be asked to include it (after eventual revisions) with the final version of your paper, and its final version will be published with the paper.

The reviewers of your paper will be asked to use the checklist as one of the factors in their evaluation. While \answerYes{} is generally preferable to \answerNo{}, it is perfectly acceptable to answer \answerNo{} provided a proper justification is given (e.g., error bars are not reported because it would be too computationally expensive'' or ``we were unable to find the license for the dataset we used''). In general, answering \answerNo{} or \answerNA{} is not grounds for rejection. While the questions are phrased in a binary way, we acknowledge that the true answer is often more nuanced, so please just use your best judgment and write a justification to elaborate. All supporting evidence can appear either in the main paper or the supplemental material, provided in appendix. If you answer \answerYes{} to a question, in the justification please point to the section(s) where related material for the question can be found.

IMPORTANT, please:
\begin{itemize}
    \item {\bf Delete this instruction block, but keep the section heading ``NeurIPS Paper Checklist"},
    \item  {\bf Keep the checklist subsection headings, questions/answers and guidelines below.}
    \item {\bf Do not modify the questions and only use the provided macros for your answers}.
\end{itemize}


\begin{enumerate}

\item {\bf Claims}
    \item[] Question: Do the main claims made in the abstract and introduction accurately reflect the paper's contributions and scope?
    \item[] Answer: \answerYes{} 
    \item[] Justification: The main ideas, contributions, and findings of the paper are clearly stated in both the Abstract and Introduction.
    \item[] Guidelines: 
    \begin{itemize}
        \item The answer \answerNA{} means that the abstract and introduction do not include the claims made in the paper.
        \item The abstract and/or introduction should clearly state the claims made, including the contributions made in the paper and important assumptions and limitations. A \answerNo{} or \answerNA{} answer to this question will not be perceived well by the reviewers. 
        \item The claims made should match theoretical and experimental results, and reflect how much the results can be expected to generalize to other settings. 
        \item It is fine to include aspirational goals as motivation as long as it is clear that these goals are not attained by the paper. 
    \end{itemize}

\item {\bf Limitations}
    \item[] Question: Does the paper discuss the limitations of the work performed by the authors?
    \item[] Answer: \answerYes{} 
    \item[] Justification: We have stated the limitations of our work at the end of our Conclusion section in the main paper.
    \item[] Guidelines:
    \begin{itemize}
        \item The answer \answerNA{} means that the paper has no limitation while the answer \answerNo{} means that the paper has limitations, but those are not discussed in the paper. 
        \item The authors are encouraged to create a separate ``Limitations'' section in their paper.
        \item The paper should point out any strong assumptions and how robust the results are to violations of these assumptions (e.g., independence assumptions, noiseless settings, model well-specification, asymptotic approximations only holding locally). The authors should reflect on how these assumptions might be violated in practice and what the implications would be.
        \item The authors should reflect on the scope of the claims made, e.g., if the approach was only tested on a few datasets or with a few runs. In general, empirical results often depend on implicit assumptions, which should be articulated.
        \item The authors should reflect on the factors that influence the performance of the approach. For example, a facial recognition algorithm may perform poorly when image resolution is low or images are taken in low lighting. Or a speech-to-text system might not be used reliably to provide closed captions for online lectures because it fails to handle technical jargon.
        \item The authors should discuss the computational efficiency of the proposed algorithms and how they scale with dataset size.
        \item If applicable, the authors should discuss possible limitations of their approach to address problems of privacy and fairness.
        \item While the authors might fear that complete honesty about limitations might be used by reviewers as grounds for rejection, a worse outcome might be that reviewers discover limitations that aren't acknowledged in the paper. The authors should use their best judgment and recognize that individual actions in favor of transparency play an important role in developing norms that preserve the integrity of the community. Reviewers will be specifically instructed to not penalize honesty concerning limitations.
    \end{itemize}

\item {\bf Theory assumptions and proofs}
    \item[] Question: For each theoretical result, does the paper provide the full set of assumptions and a complete (and correct) proof?
    \item[] Answer: \answerNA{} 
    \item[] Justification: Our paper doesn't discuss or need any theoretical results, theorems, lemmas, or proofs.
    \item[] Guidelines:
    \begin{itemize}
        \item The answer \answerNA{} means that the paper does not include theoretical results. 
        \item All the theorems, formulas, and proofs in the paper should be numbered and cross-referenced.
        \item All assumptions should be clearly stated or referenced in the statement of any theorems.
        \item The proofs can either appear in the main paper or the supplemental material, but if they appear in the supplemental material, the authors are encouraged to provide a short proof sketch to provide intuition. 
        \item Inversely, any informal proof provided in the core of the paper should be complemented by formal proofs provided in appendix or supplemental material.
        \item Theorems and Lemmas that the proof relies upon should be properly referenced. 
    \end{itemize}

    \item {\bf Experimental result reproducibility}
    \item[] Question: Does the paper fully disclose all the information needed to reproduce the main experimental results of the paper to the extent that it affects the main claims and/or conclusions of the paper (regardless of whether the code and data are provided or not)?
    \item[] Answer: \answerYes{} 
    \item[] Justification: We have carefully pointed out our method and experimental setup, including training and inference stages. The most vital parts are included in Sections 3 and 4 of the paper, and the rest are extensively discussed in the appendix.
    \item[] Guidelines:
    \begin{itemize}
        \item The answer \answerNA{} means that the paper does not include experiments.
        \item If the paper includes experiments, a \answerNo{} answer to this question will not be perceived well by the reviewers: Making the paper reproducible is important, regardless of whether the code and data are provided or not.
        \item If the contribution is a dataset and\slash or model, the authors should describe the steps taken to make their results reproducible or verifiable. 
        \item Depending on the contribution, reproducibility can be accomplished in various ways. For example, if the contribution is a novel architecture, describing the architecture fully might suffice, or if the contribution is a specific model and empirical evaluation, it may be necessary to either make it possible for others to replicate the model with the same dataset, or provide access to the model. In general. releasing code and data is often one good way to accomplish this, but reproducibility can also be provided via detailed instructions for how to replicate the results, access to a hosted model (e.g., in the case of a large language model), releasing of a model checkpoint, or other means that are appropriate to the research performed.
        \item While NeurIPS does not require releasing code, the conference does require all submissions to provide some reasonable avenue for reproducibility, which may depend on the nature of the contribution. For example
        \begin{enumerate}
            \item If the contribution is primarily a new algorithm, the paper should make it clear how to reproduce that algorithm.
            \item If the contribution is primarily a new model architecture, the paper should describe the architecture clearly and fully.
            \item If the contribution is a new model (e.g., a large language model), then there should either be a way to access this model for reproducing the results or a way to reproduce the model (e.g., with an open-source dataset or instructions for how to construct the dataset).
            \item We recognize that reproducibility may be tricky in some cases, in which case authors are welcome to describe the particular way they provide for reproducibility. In the case of closed-source models, it may be that access to the model is limited in some way (e.g., to registered users), but it should be possible for other researchers to have some path to reproducing or verifying the results.
        \end{enumerate}
    \end{itemize}

\item {\bf Open access to data and code}
    \item[] Question: Does the paper provide open access to the data and code, with sufficient instructions to faithfully reproduce the main experimental results, as described in supplemental material?
    \item[] Answer: \answerNo{} 
    \item[] Justification: Our employed data are publicly available data, as stated in the paper, and our train data curation is precisely spelt out. Our code is not yet publicly ready and included in the paper, but it will surely be accessible upon acceptance.
    \item[] Guidelines:
    \begin{itemize}
        \item The answer \answerNA{} means that paper does not include experiments requiring code.
        \item Please see the NeurIPS code and data submission guidelines (\url{https://neurips.cc/public/guides/CodeSubmissionPolicy}) for more details.
        \item While we encourage the release of code and data, we understand that this might not be possible, so \answerNo{} is an acceptable answer. Papers cannot be rejected simply for not including code, unless this is central to the contribution (e.g., for a new open-source benchmark).
        \item The instructions should contain the exact command and environment needed to run to reproduce the results. See the NeurIPS code and data submission guidelines (\url{https://neurips.cc/public/guides/CodeSubmissionPolicy}) for more details.
        \item The authors should provide instructions on data access and preparation, including how to access the raw data, preprocessed data, intermediate data, and generated data, etc.
        \item The authors should provide scripts to reproduce all experimental results for the new proposed method and baselines. If only a subset of experiments are reproducible, they should state which ones are omitted from the script and why.
        \item At submission time, to preserve anonymity, the authors should release anonymized versions (if applicable).
        \item Providing as much information as possible in supplemental material (appended to the paper) is recommended, but including URLs to data and code is permitted.
    \end{itemize}

\item {\bf Experimental setting/details}
    \item[] Question: Does the paper specify all the training and test details (e.g., data splits, hyperparameters, how they were chosen, type of optimizer) necessary to understand the results?
    \item[] Answer: \answerYes{}{} 
    \item[] Justification: We have carefully described all our train and test data in sections 3 and 4 of the main paper. Moreover, many more details are provided on that in the supplementary.
    \item[] Guidelines:
    \begin{itemize}
        \item The answer \answerNA{} means that the paper does not include experiments.
        \item The experimental setting should be presented in the core of the paper to a level of detail that is necessary to appreciate the results and make sense of them.
        \item The full details can be provided either with the code, in appendix, or as supplemental material.
    \end{itemize}

\item {\bf Experiment statistical significance}
    \item[] Question: Does the paper report error bars suitably and correctly defined or other appropriate information about the statistical significance of the experiments?
    \item[] Answer: \answerNo{} 
    \item[] Justification: It would be too computationally expensive to fine-tune VLMs and evaluate along with SAM3.
    \item[] Guidelines:
    \begin{itemize}
        \item The answer \answerNA{} means that the paper does not include experiments.
        \item The authors should answer \answerYes{} if the results are accompanied by error bars, confidence intervals, or statistical significance tests, at least for the experiments that support the main claims of the paper.
        \item The factors of variability that the error bars are capturing should be clearly stated (for example, train/test split, initialization, random drawing of some parameter, or overall run with given experimental conditions).
        \item The method for calculating the error bars should be explained (closed form formula, call to a library function, bootstrap, etc.)
        \item The assumptions made should be given (e.g., Normally distributed errors).
        \item It should be clear whether the error bar is the standard deviation or the standard error of the mean.
        \item It is OK to report 1-sigma error bars, but one should state it. The authors should preferably report a 2-sigma error bar than state that they have a 96\% CI, if the hypothesis of Normality of errors is not verified.
        \item For asymmetric distributions, the authors should be careful not to show in tables or figures symmetric error bars that would yield results that are out of range (e.g., negative error rates).
        \item If error bars are reported in tables or plots, the authors should explain in the text how they were calculated and reference the corresponding figures or tables in the text.
    \end{itemize}

\item {\bf Experiments compute resources}
    \item[] Question: For each experiment, does the paper provide sufficient information on the computer resources (type of compute workers, memory, time of execution) needed to reproduce the experiments?
    \item[] Answer: \answerYes{} 
    \item[] Justification: We mention the type of GPUs used and time of training in the supplementary.
    \item[] Guidelines:
    \begin{itemize}
        \item The answer \answerNA{} means that the paper does not include experiments.
        \item The paper should indicate the type of compute workers CPU or GPU, internal cluster, or cloud provider, including relevant memory and storage.
        \item The paper should provide the amount of compute required for each of the individual experimental runs as well as estimate the total compute. 
        \item The paper should disclose whether the full research project required more compute than the experiments reported in the paper (e.g., preliminary or failed experiments that didn't make it into the paper). 
    \end{itemize}
    
\item {\bf Code of ethics}
    \item[] Question: Does the research conducted in the paper conform, in every respect, with the NeurIPS Code of Ethics \url{https://neurips.cc/public/EthicsGuidelines}?
    \item[] Answer: \answerYes{} 
    \item[] Justification: Our research adheres to the specified Code of Ethics.
    \item[] Guidelines:
    \begin{itemize}
        \item The answer \answerNA{} means that the authors have not reviewed the NeurIPS Code of Ethics.
        \item If the authors answer \answerNo, they should explain the special circumstances that require a deviation from the Code of Ethics.
        \item The authors should make sure to preserve anonymity (e.g., if there is a special consideration due to laws or regulations in their jurisdiction).
    \end{itemize}

\item {\bf Broader impacts}
    \item[] Question: Does the paper discuss both potential positive societal impacts and negative societal impacts of the work performed?
    \item[] Answer: \answerNA{} 
    \item[] Justification: Our paper is focused on foundational research in computer vision and does not involve a special application, deployment, or technology.
    \item[] Guidelines:
    \begin{itemize}
        \item The answer \answerNA{} means that there is no societal impact of the work performed.
        \item If the authors answer \answerNA{} or \answerNo, they should explain why their work has no societal impact or why the paper does not address societal impact.
        \item Examples of negative societal impacts include potential malicious or unintended uses (e.g., disinformation, generating fake profiles, surveillance), fairness considerations (e.g., deployment of technologies that could make decisions that unfairly impact specific groups), privacy considerations, and security considerations.
        \item The conference expects that many papers will be foundational research and not tied to particular applications, let alone deployments. However, if there is a direct path to any negative applications, the authors should point it out. For example, it is legitimate to point out that an improvement in the quality of generative models could be used to generate Deepfakes for disinformation. On the other hand, it is not needed to point out that a generic algorithm for optimizing neural networks could enable people to train models that generate Deepfakes faster.
        \item The authors should consider possible harms that could arise when the technology is being used as intended and functioning correctly, harms that could arise when the technology is being used as intended but gives incorrect results, and harms following from (intentional or unintentional) misuse of the technology.
        \item If there are negative societal impacts, the authors could also discuss possible mitigation strategies (e.g., gated release of models, providing defenses in addition to attacks, mechanisms for monitoring misuse, mechanisms to monitor how a system learns from feedback over time, improving the efficiency and accessibility of ML).
    \end{itemize}
    
\item {\bf Safeguards}
    \item[] Question: Does the paper describe safeguards that have been put in place for responsible release of data or models that have a high risk for misuse (e.g., pre-trained language models, image generators, or scraped datasets)?
    \item[] Answer: \answerNA{} 
    \item[] Justification: Our paper does not raise any such risks.
    \item[] Guidelines:
    \begin{itemize}
        \item The answer \answerNA{} means that the paper poses no such risks.
        \item Released models that have a high risk for misuse or dual-use should be released with necessary safeguards to allow for controlled use of the model, for example by requiring that users adhere to usage guidelines or restrictions to access the model or implementing safety filters. 
        \item Datasets that have been scraped from the Internet could pose safety risks. The authors should describe how they avoided releasing unsafe images.
        \item We recognize that providing effective safeguards is challenging, and many papers do not require this, but we encourage authors to take this into account and make a best faith effort.
    \end{itemize}

\item {\bf Licenses for existing assets}
    \item[] Question: Are the creators or original owners of assets (e.g., code, data, models), used in the paper, properly credited and are the license and terms of use explicitly mentioned and properly respected?
    \item[] Answer: \answerYes{} 
    \item[] Justification: While we don't require any proprietary assets, all the research data, open-source VLMs that have been used in the method, and research works are properly cited in the paper and have been used according to their terms.
    \item[] Guidelines:
    \begin{itemize}
        \item The answer \answerNA{} means that the paper does not use existing assets.
        \item The authors should cite the original paper that produced the code package or dataset.
        \item The authors should state which version of the asset is used and, if possible, include a URL.
        \item The name of the license (e.g., CC-BY 4.0) should be included for each asset.
        \item For scraped data from a particular source (e.g., website), the copyright and terms of service of that source should be provided.
        \item If assets are released, the license, copyright information, and terms of use in the package should be provided. For popular datasets, \url{paperswithcode.com/datasets} has curated licenses for some datasets. Their licensing guide can help determine the license of a dataset.
        \item For existing datasets that are re-packaged, both the original license and the license of the derived asset (if it has changed) should be provided.
        \item If this information is not available online, the authors are encouraged to reach out to the asset's creators.
    \end{itemize}

\item {\bf New assets}
    \item[] Question: Are new assets introduced in the paper well documented and is the documentation provided alongside the assets?
    \item[] Answer: \answerNA{} 
    \item[] Justification: We are not releasing new assets and our work is primarily foundational research.
    \item[] Guidelines:
    \begin{itemize}
        \item The answer \answerNA{} means that the paper does not release new assets.
        \item Researchers should communicate the details of the dataset\slash code\slash model as part of their submissions via structured templates. This includes details about training, license, limitations, etc. 
        \item The paper should discuss whether and how consent was obtained from people whose asset is used.
        \item At submission time, remember to anonymize your assets (if applicable). You can either create an anonymized URL or include an anonymized zip file.
    \end{itemize}

\item {\bf Crowdsourcing and research with human subjects}
    \item[] Question: For crowdsourcing experiments and research with human subjects, does the paper include the full text of instructions given to participants and screenshots, if applicable, as well as details about compensation (if any)? 
    \item[] Answer: \answerNA{} 
    \item[] Justification: Our paper does not involve crowdsourcing nor research with human subjects.
    \item[] Guidelines:
    \begin{itemize}
        \item The answer \answerNA{} means that the paper does not involve crowdsourcing nor research with human subjects.
        \item Including this information in the supplemental material is fine, but if the main contribution of the paper involves human subjects, then as much detail as possible should be included in the main paper. 
        \item According to the NeurIPS Code of Ethics, workers involved in data collection, curation, or other labor should be paid at least the minimum wage in the country of the data collector. 
    \end{itemize}

\item {\bf Institutional review board (IRB) approvals or equivalent for research with human subjects}
    \item[] Question: Does the paper describe potential risks incurred by study participants, whether such risks were disclosed to the subjects, and whether Institutional Review Board (IRB) approvals (or an equivalent approval/review based on the requirements of your country or institution) were obtained?
    \item[] Answer: \answerNA 
    \item[] Justification: The paper does not involve crowdsourcing nor research with human subjects.
    \item[] Guidelines:
    \begin{itemize}
        \item The answer \answerNA{} means that the paper does not involve crowdsourcing nor research with human subjects.
        \item Depending on the country in which research is conducted, IRB approval (or equivalent) may be required for any human subjects research. If you obtained IRB approval, you should clearly state this in the paper. 
        \item We recognize that the procedures for this may vary significantly between institutions and locations, and we expect authors to adhere to the NeurIPS Code of Ethics and the guidelines for their institution. 
        \item For initial submissions, do not include any information that would break anonymity (if applicable), such as the institution conducting the review.
    \end{itemize}

\item {\bf Declaration of LLM usage}
    \item[] Question: Does the paper describe the usage of LLMs if it is an important, original, or non-standard component of the core methods in this research? Note that if the LLM is used only for writing, editing, or formatting purposes and does \emph{not} impact the core methodology, scientific rigor, or originality of the research, declaration is not required.
    \item[] Answer: \answerYes{} 
    \item[] Justification: With our work exploiting open-source VLMs for our research, we have carefully introduced the used models and our way of using them in the paper. Moreover, the detailed exact prompt used in our work to prompt the targeted VLM is provided in the appendix.
    \item[] Guidelines:
    \begin{itemize}
        \item The answer \answerNA{} means that the core method development in this research does not involve LLMs as any important, original, or non-standard components.
        \item Please refer to our LLM policy in the NeurIPS handbook for what should or should not be described.
    \end{itemize}

\end{enumerate}

\end{document}